\pdfoutput=1
\documentclass[10pt,twocolumn,letterpaper]{article}

\usepackage[pagenumbers]{cvpr}

\usepackage{float}
\usepackage{capt-of}
\usepackage[colaction]{multicol}
\usepackage{needspace}
\makeatletter
\newsavebox\supp@widebox

\makeatother

\usepackage[utf8]{inputenc}
\usepackage[T1]{fontenc}
\usepackage{amsfonts}
\usepackage{nicefrac}
\usepackage{microtype}
\usepackage{colortbl}
\usepackage{multirow}
\definecolor{bestblue}{RGB}{229,238,248}
\definecolor{secondblue}{RGB}{244,248,253}
\newcommand{\best}[1]{\cellcolor{bestblue}#1}
\newcommand{\secondbest}[1]{\cellcolor{secondblue}#1}

\newcommand{\supp}[2][Appendix]{#1~\ref{#2}}

\definecolor{cvprblue}{rgb}{0.21,0.49,0.74}
\usepackage[pagebackref,breaklinks,colorlinks,allcolors=cvprblue]{hyperref}

\def\confName{CVPR}
\def\confYear{2027}

\title{LaWM: Learning Physics from Real Video with Least Action World Models for Long-Horizon Consistency}

\author{Qixin Xiao \qquad Maani Ghaffari\\
University of Michigan\\
{\tt\small \{qxiaocs, maanigj\}@umich.edu}
}

\begin{document}
\twocolumn[{%
\renewcommand\twocolumn[1][]{#1}%
\maketitle
\vspace{-1.2em}
\begin{center}
    \centering
    \captionsetup{type=figure}
    \includegraphics[width=\textwidth]{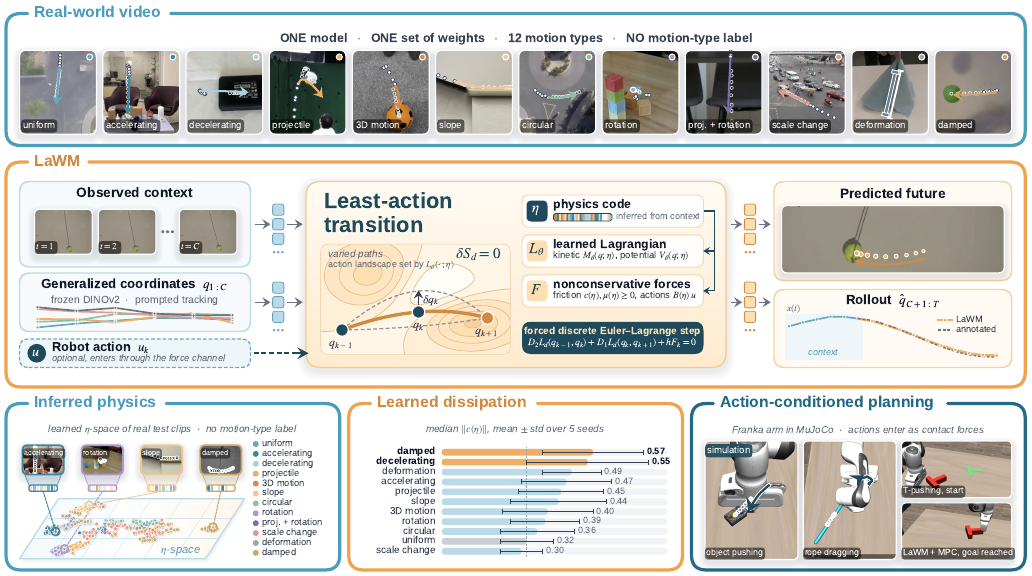}
    \caption{Overview of LaWM. \textbf{Top:} real test clips of twelve motion types, all predicted by one model with one set of weights and no motion-type label. White marks the observed context and color marks the LaWM rollout. \textbf{Middle:} a frozen DINOv2 backbone with first-frame prompted tracking reads the generalized coordinates $q_{1:C}$ from the context frames. A context encoder infers a physics code $\eta$, which conditions a learned Lagrangian $L_\theta$ and nonconservative forces $F$. Each future state is the solution of the forced discrete Euler--Lagrange equation, applied recursively to roll out the future. The action landscape is schematic. \textbf{Bottom:} 2-D t-SNE embedding of $\eta$ on real test clips, with selected clips showing $\eta$ in its top-8 principal components (left). Median inferred viscous coefficient $\|c(\eta)\|$ per motion type, mean and standard deviation over five seeds (center). Action-conditioned planning with a Franka arm in MuJoCo simulation (right).}
    \label{fig:main_architecture}
\end{center}
\vspace{0.6em}
}]
\begin{abstract}
Learning world models that predict how the physical world evolves from visual observations is central to embodied AI and to robotic planning. Visual world models advance learned features with unconstrained transition networks, so physical laws do not govern the transition. Physics-grounded approaches either take physics as given through simulators, losses, or guidance, or learn it with structured networks trained on states or simulated renderings of one physical system at a time. We present Least Action World Models (LaWM), a unified visual world model whose transition rule is the Principle of Least Action over learnable physical properties. (i) LaWM operates directly on real video, where a frozen visual foundation model reads object-centric generalized coordinates and a learned variational integrator advances them without ground-truth states or simulators. (ii) The Lagrangian and the nonconservative forces are inferred from the observed context through the discrete Lagrange--d'Alembert principle, so that a single model covers conservative, damped, and driven motion. (iii) External forces enter the update as an explicit channel, which extends LaWM to action-conditioned prediction. On real videos of twelve motion types, LaWM outperforms complete perception and prediction systems built on black-box, continuous-time, Lagrangian, and variational-integrator transitions in prediction accuracy and in the rate of diverging rollouts, and retains this advantage over $42$-frame rollouts. Given the same coordinates, LaWM matches the accuracy of black-box transitions while its Lagrangian and forces remain explicit, and adding learned forces to a conservative variational integrator most improves the prediction of dissipative motion.
\end{abstract}

\section{Introduction}

A world model predicts how the environment evolves, and this capability underlies model-based reinforcement learning and robotic planning~\citep{hafner2025dreamerv3,zhou2025dinowm,assran2025vjepa2}.
For embodied agents, the observations are images, while the evolution of a scene is governed by a small number of physical quantities, such as the position, momentum, and energy of the objects in it.
A visual world model should therefore recover these quantities from video and advance them in a way that respects the laws they obey.

Current approaches fall short of this goal in two ways.
Visual world models encode observations into learned features and advance them with unconstrained transition networks~\citep{hafner2025dreamerv3,zhou2025dinowm,assran2025vjepa2}.
Physical laws do not govern these transitions, and nothing prevents a rollout from drifting away from physically admissible motion.
Large video generators share this limitation, and recent benchmarks built from real experiments show that visually convincing videos frequently violate basic mechanics~\citep{motamed2025physicsiq,zhang2025morpheus,li2025pisa}.
Physics-grounded approaches take one of two routes.
The first treats physics as given, coupling generation with physics simulators~\citep{liu2024physgen,xie2024physgaussian,zhang2024physdreamer}, with prompt guidance from a language model~\citep{xue2025phyt2v}, or with auxiliary depth and keypoint losses~\citep{shang2025roboscape}.
The second learns physics with structured networks such as Hamiltonian and Lagrangian networks~\citep{greydanus2019hnn,cranmer2020lnn,lutter2019delan}, variational integrator networks~\citep{saemundsson2020vin,havens2021fvin}, and neural Newtonian dynamics~\citep{yuan2025newtongen}.
These models are trained on low-dimensional states or simulated renderings of one physical system at a time, and most of them assume energy conservation, which real motion, being dissipative or driven, violates.

We present Least Action World Models (LaWM), a unified visual world model in which the transition rule is the Principle of Least Action over learnable physical properties (Fig.~\ref{fig:main_architecture}).
A frozen visual foundation model, prompted with the target object in the first frame, reads object-centric generalized coordinates from raw video.
A context encoder infers a physics code from the observed frames, and this code conditions a learned discrete Lagrangian together with dissipative and external generalized forces.
The next state is the solution of the forced discrete Euler--Lagrange equation that follows from the discrete Lagrange--d'Alembert principle~\citep{marsden2001discrete}.
Physics therefore does not score or correct a predicted trajectory but defines the transition, a variational integrator whose energy changes through the learned forces rather than through numerical damping.
Unlike prior variational integrator networks, which learn one Lagrangian per physical system, LaWM infers the Lagrangian and the forces per clip, so a single model covers twelve real motion types and conservative, damped, and driven motion.

We evaluate LaWM on real videos of twelve motion types collected from public datasets, without any simulated training data, and the results support three findings.
First, LaWM is an effective world model on real video.
Against complete perception and prediction systems that combine a frozen foundation model readout with black-box, continuous-time, Lagrangian, and variational-integrator transitions, it reduces the trajectory error of the strongest baseline by $40\%$ and the rate of diverging rollouts from $16.2\%$ to $5.7\%$, and the advantage persists with a quarter of the training data, beyond the training horizon, and on motion types held out during training.
Controlled comparisons attribute most of this gain to accurate generalized coordinates.
Second, the physical structure costs no accuracy.
Given the same coordinates, the forced variational integrator matches the accuracy of black-box transitions and is significantly more accurate than a variational integrator without learned forces and context inference.
Third, the explicit components carry physical meaning.
Learned nonconservative forces reduce the error of a conservative variational integrator most on damped oscillation and deformation, where energy is dissipated, the learned potential correlates with measured slope angles and pendulum lengths without being fitted to them, and with the same coordinates the physics-structured transitions attain lower mean errors than the black-box transitions on unseen motion types.
As a complete system, LaWM also leads on $42$-frame rollouts; with the same coordinates, the best black-box transitions are more accurate under direct training, whereas LaWM attains the lowest error under a validation-selected horizon curriculum.
Simulated manipulation scenes demonstrate that the same model accepts robot actions through its force channel.

Our contributions are threefold.
\begin{itemize}
    \item[(i)] A world model that operates directly on real video, in which a frozen visual foundation model reads object-centric generalized coordinates and a learned variational integrator advances them without ground-truth states or simulators.
    \item[(ii)] Explicit learned physics, in which the Lagrangian and the nonconservative forces are inferred from the context through the discrete Lagrange--d'Alembert principle, so that one model covers conservative, damped, and driven motion and matches black-box accuracy.
    \item[(iii)] An explicit control interface, since external forces enter the update as a separate channel, which we demonstrate in simulated manipulation.
\end{itemize}

\section{Related Work}
\label{sec:related_work}

\begin{figure*}[!t]
    \centering
    \includegraphics[width=\textwidth]{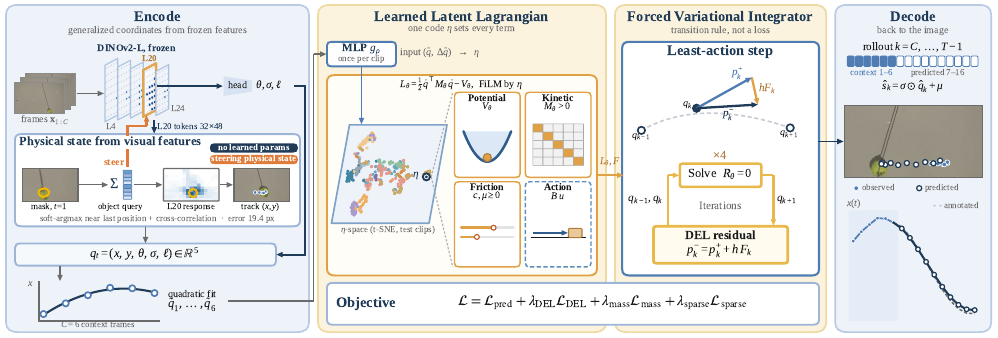}
    \caption{The LaWM pipeline. \textbf{Encode:} each frame is encoded by a frozen DINOv2-L backbone. A mask on the first frame defines an object query that steers the readout of the frozen layer-20 features: its cosine response selects where the object is read out, without modifying the backbone or learning parameters. Orientation, size, and shape are read from the same features by a light head. The context coordinates are smoothed by a quadratic fit. \textbf{Learned Latent Lagrangian:} an MLP maps the smoothed context to a physics code $\eta$, shown in its t-SNE embedding over the test clips. The code sets the potential $V_\theta$, the positive mass $M_\theta$, the nonnegative friction coefficients $c$ and $\mu$, and the action input $B\,u$. \textbf{Forced Variational Integrator:} the forced discrete Euler--Lagrange residual expresses the momentum balance $p_k^-=p_k^++hF_k$ at $q_k$, and $J=4$ mass-preconditioned iterations solve it for $q_{k+1}$. \textbf{Decode:} the rollout over frames $C{+}1,\dots,T$ is mapped back to image coordinates. Images, the layer-20 response, the track, the context fit, the $\eta$ embedding, and the rollout are real outputs for a test clip. The layer activations, the momentum diagram, and the term icons are schematic.}
    \label{fig:method}
\vspace{-3mm}
\end{figure*}

\subsection{Visual World Models}
\label{sec:rw_world_models}

Latent world models such as PlaNet and Dreamer learn compact states for prediction and planning~\citep{hafner2019planet,hafner2020dreamer,hafner2025dreamerv3}, and deep Koopman models seek coordinates with linear dynamics~\citep{lusch2018deepkoopman}.
V-JEPA learns video encoders by predicting in their own feature space~\citep{bardes2024vjepa}, V-JEPA 2 plans with an action-conditioned predictor over its frozen encoder~\citep{assran2025vjepa2}, and DINO-WM plans with a predictor over frozen DINOv2 patch features~\citep{zhou2025dinowm}.
Video generators approach prediction through synthesis~\citep{blattmann2023videoldm,openai2024sora,yang2025cogvideox}, yet benchmarks built on real experiments show that their outputs often violate basic mechanics~\citep{motamed2025physicsiq,zhang2025morpheus,li2025pisa}.
In these models no physical law governs the prediction, whereas the transition of LaWM is the forced discrete Euler--Lagrange equation of a learned Lagrangian.

\subsection{Physics-Grounded Video Generation}
\label{sec:rw_physics_video}

Other methods introduce physics into generation as given knowledge, through rigid-body or material point method simulators~\citep{liu2024physgen,xie2024physgaussian,zhang2024physdreamer}, prompt refinement with a language model~\citep{xue2025phyt2v}, or auxiliary depth and keypoint-dynamics objectives~\citep{shang2025roboscape}.
NewtonGen guides a video generator with neural Newtonian dynamics learned per motion type from simulated videos~\citep{yuan2025newtongen}.
Physics-informed neural networks impose prescribed differential equations, possibly with unknown coefficients, as losses~\citep{raissi2019physicsinformed,karniadakis2021physicsinformed}.
LaWM instead infers the energy landscape and the forces of each scene from real video with one model for all motion types and uses them as the transition rule itself.

\subsection{Learning Structured Dynamics}
\label{sec:rw_structured}

Neural ODEs learn free-form continuous-time dynamics~\citep{chen2018neuralode}, and Hamiltonian and Lagrangian networks restrict them to energy-based mechanics~\citep{greydanus2019hnn,cranmer2020lnn,lutter2019delan}.
Discrete mechanics derives symplectic variational integrators with good long-time behavior from a discrete action principle~\citep{marsden2001discrete,hairer2006geometric}.
Variational Integrator Networks learn a discrete Lagrangian, including in the latent space of an image autoencoder~\citep{saemundsson2020vin}, Discrete Lagrangian Neural Networks additionally discover symmetries~\citep{lishkova2023dlnn}, and forced variants add dissipation and control through the discrete Lagrange--d'Alembert principle, on Lie groups, and from position data of measured or rendered systems~\citep{havens2021fvin,duruisseaux2023liefvin,hansen2025dflnn}.
LaWM follows this line, but these models learn one physical system at a time from states or from images of a single controlled scene.
LaWM differs in three respects.
It reads generalized coordinates from real video with a frozen foundation model and prompted tracking.
It conditions the Lagrangian and the forces on a physics code inferred from the context, so that one model covers twelve motion types without motion labels.
It is evaluated as a complete perception and prediction system against black-box, continuous-time, Lagrangian, and variational-integrator transitions on the same data.

\subsection{Physical Inference from Real Video}
\label{sec:rw_inference}

Physics~101 learns object properties from unlabeled videos~\citep{wu2016physics101}, and recent work estimates the parameters of known equations from a single video~\citep{delfys75}, benchmarks the recovery of parameters and equations against measured ground truth~\citep{iris}, and reads dynamic properties from video foundation models~\citep{physvid}.
These methods assume a known equation family or target a specific property.
LaWM requires no governing equation and evaluates the inferred physics by the accuracy of the resulting rollout, and the real videos released by these works form a large part of our data.

\section{Method}
\label{sec:method}

LaWM predicts the future of an observed object by integrating learned mechanics rather than regressing the next state.
A frozen visual foundation model reads the generalized coordinates of the object from each frame (Sec.~\ref{sec:encode}).
A context encoder infers a physics code from the observed frames (Sec.~\ref{sec:context}), which conditions a learned Lagrangian (Sec.~\ref{sec:lagrangian}) and a set of nonconservative forces (Sec.~\ref{sec:forces}).
The next state is obtained by solving the forced discrete Euler--Lagrange equation that these quantities define (Sec.~\ref{sec:del}).
The Principle of Least Action, extended to forced systems by the Lagrange--d'Alembert principle, is therefore the transition rule itself and not an auxiliary objective.
Fig.~\ref{fig:method} details the pipeline.
A single model with one set of weights is trained on all motion types and receives no motion label.

\subsection{Problem Setup}
\label{sec:setup}

A clip $x_{1:T}$ shows one moving object.
The state of the object in frame $t$ is
\begin{equation}
    s_t = (x_t,\, y_t,\, \theta_t,\, \sigma_t,\, \ell_t)\in\mathbb{R}^{5},
\end{equation}
consisting of the image position, the in-plane orientation, the apparent size, and a shape coordinate given by the length of the major axis.
The model observes the first $C$ frames and predicts the states of the remaining $T-C$ frames.
Ground-truth or pseudo-label states serve only as training and evaluation targets and are never given to the model as input.
In the standard setting $T=16$ and $C=6$.

\subsection{Generalized Coordinates from Video}
\label{sec:encode}

\paragraph{Prompted tracking.}
Each frame is resized to $448\times672$ and encoded independently by a frozen DINOv2-L backbone~\citep{oquab2023dinov2}.
We use the layer-20 patch tokens on their native $32\times48$ grid.
The object to be modeled is specified once, by a mask in the first frame.
The mask-weighted mean of the first-frame tokens, normalized to unit length, forms an object query $\tau$.
The query steers the readout of the frozen features without modifying the backbone.
In frame $k$, the cosine similarity between $\tau$ and every layer-20 token gives a response map $S_k$, which selects where the object is read out.
The position estimate $p_k^{\mathrm{loc}}$ is the soft-argmax of $S_k/\gamma$ within a disc around the peak of $S_k$, where the peak is searched within $48$ pixels of the previous estimate and $\gamma=0.03$.
A second estimate propagates the previous position by the displacement $d_k$ that maximizes the cross-correlation between a Gaussian-weighted feature patch around $p_{k-1}$ in frame $k-1$ and the features of frame $k$.
The two estimates are fused as
\begin{equation}
    p_k = \alpha\,(p_{k-1}+d_k) + (1-\alpha)\,p_k^{\mathrm{loc}},
    \label{eq:track}
\end{equation}
with $\alpha=0.75$.
Applying the same procedure to the first frame yields a constant offset to the known first-frame position, which is subtracted from all estimates.
The tracker has no trainable parameters, and its hyperparameters were selected on the validation split.
Orientation, size, and shape are read from the same frozen features by a lightweight attention-pooling head trained on the training frames.

\paragraph{Coordinates.}
The generalized coordinates are the perceived states normalized by the mean and standard deviation of the training set, $q_t=(\hat s_t-\mu)/\sigma\in\mathbb{R}^5$.

\paragraph{Context smoothing.}
Errors in the per-frame coordinates matter more for a mechanical model than for a direct predictor, because an error in the initial velocity propagates through the entire rollout.
The $C$ context coordinates are therefore replaced by their least-squares quadratic fit in time, $\tilde q_{1:C} = P\, q_{1:C}$ with the fixed projection $P = V(V^\top V)^{-1}V^\top$, $V_{ij}=i^{\,j}$, $j\in\{0,1,2\}$, and the same smoothing is applied to every baseline.
The rollout starts from the last two smoothed states, $\tilde q_{C-1}$ and $\tilde q_C$.

\subsection{Inferring the Physics from Context}
\label{sec:context}

Free fall, sliding with friction, and damped oscillation differ in their energy landscape and their dissipation, and one model must read both from the observed frames.
A multilayer perceptron $g_\rho$ maps the smoothed context and its frame-to-frame differences to a physics code
\begin{equation}
    \eta = g_\rho\big(\tilde q_{1:C},\ \tilde q_{2:C}-\tilde q_{1:C-1}\big)\in\mathbb{R}^{32}.
    \label{eq:eta}
\end{equation}
The code is computed once per clip and held fixed during the rollout.
It receives no supervision beyond the prediction objective and no motion label.

\subsection{Learned Lagrangian}
\label{sec:lagrangian}

The conservative part of the dynamics is a Lagrangian conditioned on $\eta$,
\begin{equation}
    L_\theta(q,\dot q;\eta) = \tfrac12\, \dot q^\top M_\theta(q,\eta)\, \dot q \;-\; V_\theta(q,\eta).
    \label{eq:lagrangian}
\end{equation}
The potential $V_\theta$ is a multilayer perceptron with scalar output.
The mass matrix $M_\theta=\mathrm{diag}\big(\mathrm{softplus}(m_\theta(q,\eta))+m_{\min}\big)$ is diagonal and positive, so the kinetic energy is nonnegative, and it is initialized to the identity.
Both networks receive $\eta$ through feature-wise linear modulation (FiLM), so a single network represents a family of energy landscapes indexed by the physics code.

Video is sampled at discrete times, so the model is defined directly through a discrete Lagrangian on consecutive states.
With step size $h$ equal to one frame,
\begin{equation}
    L_d(q_k,q_{k+1};\eta) = h\, L_\theta\Big(\tfrac{q_k+q_{k+1}}{2},\ \tfrac{q_{k+1}-q_k}{h};\ \eta\Big),
    \label{eq:discrete_lagrangian}
\end{equation}
and the discrete action of a trajectory is $\mathcal S_d=\sum_k L_d(q_k,q_{k+1};\eta)$.

\subsection{Nonconservative Forces}
\label{sec:forces}

A Lagrangian alone describes only systems that conserve energy, whereas most real motion loses energy or is driven.
Forces that cannot be derived from a potential enter through the Lagrange--d'Alembert principle as a generalized force $F_k$ on step $k$.

\paragraph{Dissipation.}
Energy loss is modeled by the two standard friction laws,
\begin{equation}
    \begin{gathered}
    F^{\mathrm{dis}}_k = -\,c(\eta)\odot v_k \;-\; \mu(\eta)\odot \tanh(v_k/\varepsilon),\\
    v_k=(q_k-q_{k-1})/h,
    \end{gathered}
    \label{eq:dissipation}
\end{equation}
where the first term is viscous damping and the second is Coulomb friction, with $\tanh(\cdot/\varepsilon)$, $\varepsilon=0.02$, as a smooth sign function.
The coefficients $c(\eta),\mu(\eta)\in\mathbb{R}^5_{\ge0}$ depend only on the physics code and not on the state.
They describe the setting of a clip, such as the damping of a pendulum, and cannot act as a state-dependent correction.
Because the coefficients are nonnegative, the power of the dissipative force is $F^{\mathrm{dis}\top}_k v_k\le 0$, so this force can only remove energy.
Each coefficient is a softplus output scaled by a learned gate initialized to $0.01$, $c(\eta)=\mathrm{relu}(g_c)\,\mathrm{softplus}(a_c(\eta))$ and likewise for $\mu$, so training starts close to the conservative model.

\paragraph{External force.}
When an action or a measured contact force $u_k$ is available, it enters as
\begin{equation}
    F^{\mathrm{ext}}_k = B(\eta)\,u_k ,
    \label{eq:external}
\end{equation}
with a learned input matrix $B(\eta)$ initialized to zero.
This is the only term that can inject energy and the interface for control inputs, and it is inactive on passive video.

The total generalized force is $F_k=F^{\mathrm{dis}}_k+F^{\mathrm{ext}}_k$.

\subsection{Forced Discrete Euler--Lagrange Transition}
\label{sec:del}

\paragraph{Transition equation.}
The discrete Lagrange--d'Alembert principle requires that the variation of the discrete action plus the virtual work of the forces vanishes for every interior state.
For state $q_k$ this yields the forced discrete Euler--Lagrange (DEL) equation~\citep{marsden2001discrete}
\begin{equation}
    \begin{split}
    R_\theta(q_{k-1},q_k,q_{k+1};\eta) ={}& D_2L_d(q_{k-1},q_k;\eta)\\
    &+D_1L_d(q_k,q_{k+1};\eta)+h\,F_k = 0,
    \end{split}
    \label{eq:forced_del}
\end{equation}
where $D_1$ and $D_2$ denote derivatives with respect to the first and second argument.
With the discrete momenta $p_k^+=D_2L_d(q_{k-1},q_k)$ and $p_k^-=-D_1L_d(q_k,q_{k+1})$, Eq.~\eqref{eq:forced_del} reads $p_k^-=p_k^++h\,F_k$, a discrete form of Newton's second law in which the momentum leaving a state equals the momentum arriving plus the impulse of the forces.
The equation is implicit in $q_{k+1}$, and the next state is defined as its solution.
Without forces, the resulting map is a variational integrator that preserves a discrete symplectic form, and with forces its energy follows the dissipation of the forces without artificial numerical damping~\citep{marsden2001discrete}.
A direct predictor $q_{k+1}=f(q_k,q_{k-1})$ carries no such structure.

\paragraph{Solver.}
Given $q_{k-1}$ and $q_k$, Eq.~\eqref{eq:forced_del} is solved for $q'$ by $J=4$ differentiable, mass-preconditioned iterations starting from the constant-velocity guess,
\begin{equation}
    \begin{gathered}
    q'^{(0)} = 2q_k-q_{k-1},\\
    \Delta^{(j)} = M_\theta\big(\tfrac12(q_k+q'^{(j)}),\eta\big)^{-1} R_\theta\big(q_{k-1},q_k,q'^{(j)};\eta\big),\\
    q'^{(j+1)} = q'^{(j)} + \min\!\Big(1,\ \frac{\Delta_{\max}}{\|\Delta^{(j)}\|}\Big)\,\Delta^{(j)},
    \end{gathered}
    \label{eq:solver}
\end{equation}
and $q_{k+1}=q'^{(J)}$.
Since $\partial R_\theta/\partial q' \approx -M_\theta/h$, preconditioning by the inverse mass is a Newton-type step.
The step cap $\Delta_{\max}=2$ in normalized units prevents a single poorly conditioned step from destabilizing the rollout.
The force uses the explicit velocity $v_k$, so it is evaluated once per step, and gradients are propagated through all iterations.

\paragraph{Rollout.}
Starting from $(\tilde q_{C-1},\tilde q_C)$, the solve of Eq.~\eqref{eq:solver} is applied recursively for $k=C,\dots,T-1$.
All motion types share the same transition and differ only in $\eta$.
Predicted states are mapped back to image units by $\hat s_k=\sigma\odot\hat q_k+\mu$.

\subsection{Training and Inference}
\label{sec:training}

All losses are computed in normalized coordinates.
The prediction loss compares the rollout with the target states,
\begin{equation}
    \mathcal L_{\mathrm{pred}}=\frac{1}{T-C}\sum_{k=C+1}^{T}\big\|\hat q_k-q^{\star}_k\big\|^2 ,
\end{equation}
where $q^{\star}_k$ is the normalized target state.
The DEL loss requires the target trajectory itself to satisfy the learned mechanics.
At a random interior step $k$ it penalizes the mass-scaled residual,
\begin{equation}
    \mathcal L_{\mathrm{DEL}}=\big\|M_\theta^{-1}R_\theta(q^{\star}_{k-1},q^{\star}_k,q^{\star}_{k+1};\eta)\big\|^2 ,
\end{equation}
which ties the Lagrangian and the forces to the observed motion and not only to the rollout.
Two regularizers complete the objective.
$\mathcal L_{\mathrm{mass}}=\overline{(\log m_\theta)^2}$ fixes the scale ambiguity between mass and energy, and $\mathcal L_{\mathrm{sparse}}=\overline{c(\eta)}+\overline{\mu(\eta)}$ keeps the friction coefficients small unless a clip requires them.
The full objective is
\begin{equation}
    \begin{split}
    \mathcal L={}&
    \mathcal L_{\mathrm{pred}}
    +\lambda_{\mathrm{DEL}}\mathcal L_{\mathrm{DEL}}\\
    &+\lambda_{\mathrm{mass}}\mathcal L_{\mathrm{mass}}
    +\lambda_{\mathrm{sparse}}\mathcal L_{\mathrm{sparse}},
    \end{split}
    \label{eq:objective}
\end{equation}
with $\lambda_{\mathrm{DEL}}=0.1$ and $\lambda_{\mathrm{mass}}=\lambda_{\mathrm{sparse}}=10^{-3}$.
The context of every training clip is taken from the tracker, as at test time, while the targets are the annotated states.
Clips are sampled so that all motion types appear equally often, and the motion label is used only for this sampling.
The checkpoint with the lowest validation error is selected.

At test time, LaWM tracks the object through the $C$ context frames, smooths the coordinates, infers $\eta$, rolls out the forced DEL transition, and maps the predicted coordinates back to image units.
No motion label, future frame, or annotated state is used.

\section{Experiments}
\label{sec:experiments}

\begin{table*}[!t]
\centering
\small
\setlength{\tabcolsep}{4pt}
\caption{Comparison of complete perception and prediction systems on real video. Baselines combine conventional perception (a ridge readout of global DINOv2 features, corrected with the same first-frame information as ours) with their own transition. DINO-WM~\citep{zhou2025dinowm} predicts frozen DINOv2 patch features and is decoded to positions by our prompted tracker on the predicted features (the global readout is worse, \supp{app:dinowm}); it is not evaluated on the seven Falling Objects clips. LaWM combines prompted tracking with the forced variational integrator. Normalized $xy$ trajectory error over $10$ predicted frames, lower is better, averaged over $5$ seeds. Mean is reported with the standard deviation over seeds. Mean, Median, and Div.\ are computed over the $554$ test clips of the other eleven motion types. Div.\ is the percentage of clips with error above $1$. The best and second-best results in each row are highlighted in dark and light blue, and our results are in bold. $^\dagger$One of five Neural ODE seeds diverges numerically on a single uniform-motion clip.}
\label{tab:main}
\resizebox{\ifdim\width>\textwidth\textwidth\else\width\fi}{!}{%
\begin{tabular}{lccccccccc}
\toprule
& \multicolumn{3}{c}{Black-box} & Continuous & \multicolumn{3}{c}{Physics-structured} & Feature WM & \\
\cmidrule(lr){2-4}\cmidrule(lr){5-5}\cmidrule(lr){6-8}\cmidrule(lr){9-9}
Motion & MLP & GRU & Transformer & Neural ODE & LNN & VIN & FVIN & DINO-WM & LaWM (ours) \\
\midrule
Uniform & 1.221 & 0.944 & 0.711 & $>\!10^3$$^\dagger$ & 0.792 & 0.841 & 0.666 & \secondbest{0.602} & \bfseries \best{0.383} \\
Acceleration & 0.525 & 0.503 & 0.497 & 0.540 & 0.598 & 0.535 & \secondbest{0.484} & 0.862 & \bfseries \best{0.308} \\
Deceleration & 0.361 & 0.326 & 0.300 & 0.356 & 0.319 & 0.475 & 0.326 & \best{0.225} & \bfseries \secondbest{0.296} \\
Parabolic & 0.619 & 0.575 & 0.579 & 0.637 & 0.577 & 0.697 & 0.748 & \secondbest{0.557} & \bfseries \best{0.374} \\
Parabolic + rotation & 0.725 & 0.757 & 0.802 & 0.728 & 0.725 & 0.762 & \secondbest{0.594} & -- & \bfseries \best{0.328} \\
3D & 0.482 & 0.479 & 0.515 & 0.476 & 0.454 & 0.541 & 0.522 & \secondbest{0.245} & \bfseries \best{0.214} \\
Size change & 0.934 & 0.943 & 0.918 & 0.932 & 0.941 & 1.044 & 0.985 & \secondbest{0.891} & \bfseries \best{0.549} \\
Slope & 0.434 & 0.403 & \secondbest{0.400} & 0.437 & 0.525 & 0.411 & 0.456 & 1.310 & \bfseries \best{0.259} \\
Circular & 0.929 & 0.884 & \secondbest{0.852} & 0.925 & 1.010 & 0.871 & 0.936 & 1.085 & \bfseries \best{0.389} \\
Rotation & 0.394 & 0.419 & 0.424 & 0.392 & 0.387 & 0.461 & \secondbest{0.362} & 0.365 & \bfseries \best{0.344} \\
Damped oscillation & 0.686 & 0.659 & 0.654 & 0.674 & 0.669 & 0.774 & 0.703 & \secondbest{0.498} & \bfseries \best{0.375} \\
Deformation & 0.277 & 0.259 & 0.258 & 0.307 & 0.244 & 0.363 & \best{0.185} & \secondbest{0.220} & \bfseries 0.236 \\
\midrule
Mean & 0.696$\pm$0.055 & 0.648$\pm$0.013 & \secondbest{0.615$\pm$0.034} & 614$\pm$1227$^\dagger$ & 0.657$\pm$0.011 & 0.702$\pm$0.028 & 0.647$\pm$0.011 & 0.681$\pm$0.003 & \bfseries \best{0.368$\pm$0.012} \\
Median & 0.571 & 0.544 & \secondbest{0.514} & 0.575 & 0.581 & 0.599 & 0.589 & 0.590 & \bfseries \best{0.266} \\
Div.\ (\%) & 18.4 & 17.8 & \secondbest{16.2} & 17.9 & 16.9 & 20.7 & 17.1 & 24.0 & \bfseries \best{5.7} \\
\bottomrule
\end{tabular}}
\vspace{-3mm}
\end{table*}

\subsection{Experimental Setup}
\label{sec:exp_setup}

\paragraph{Data.}
All training and test clips are real videos drawn from sixteen public datasets~\citep{milan2016mot16,robicquet2016sdd,li2025pisa,motamed2025physicsiq,zhang2025morpheus,delfys75,physvid,dabral2021gravicap,sun2020tracknetv2,rozumnyi2020tbd3d,kotera2020restoration,iris,wen2020uadetrac,wu2016physics101,gsphysics,jiang2025phystwin}, covering the twelve motion types of NewtonGen~\citep{yuan2025newtongen}: uniform, accelerated, decelerated, parabolic, parabolic with rotation, 3D, size-changing, slope, circular, rotational, damped oscillatory, and deformable motion.
The set contains $1{,}616$ clips ($819$ train, $236$ validation, $561$ test) of $16$ frames, split by recording event.
Object states from annotations or SAM2~\citep{ravi2024sam2} masks serve only as targets (\supp{app:datasets}).

\paragraph{Perception.}
The baselines use the standard readout of a frozen foundation model, a ridge regression from mean-pooled DINOv2-L features~\citep{oquab2023dinov2} to the state, corrected by its first-frame offset, which lowers its localization error from $89.5$ to $61.0$ pixels, against $19.4$ pixels for our tracker.

\paragraph{Baselines.}
We compare against seven transition rules from three families: a second-order MLP, $q_{k+1}=2q_k-q_{k-1}+f(q_k,q_{k-1})$, a GRU, and a Transformer over the full history (black-box); a Neural ODE~\citep{chen2018neuralode} (continuous-time); and an LNN~\citep{cranmer2020lnn}, a VIN~\citep{saemundsson2020vin}, and an FVIN~\citep{havens2021fvin} (physics-structured).
All share the context, smoothing, loss, optimizer, and budget of LaWM, and each is one model for all motion types (more baselines in \supp{app:extra_baselines}).

\paragraph{Metrics and statistics.}
The main metric is the normalized root mean squared error (NRMSE) of the predicted $(x,y)$ trajectory, with each coordinate normalized by its training-set standard deviation.
We report the mean and median over test clips and the divergence rate (Div.), the percentage of clips with NRMSE above $1$, over $5$ seeds.
A difference is significant at the seed level when it exceeds twice the sum of the standard deviations, and we also apply a paired bootstrap over test clips (\supp{app:bootstrap}).
Metrics and tests were fixed in advance, and the test set was not used for any selection (details in \supp{app:implementation}).

\subsection{Comparison of Complete Systems}
\label{sec:exp_main}

\begin{figure}[!htb]
\centering
\includegraphics[width=\columnwidth]{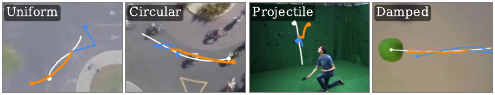}
\caption{Predictions on real test clips, with the annotated trajectory (white), LaWM (orange), and the conventional system with the Transformer transition (blue). Each clip is chosen by a fixed rule, as the clip with LaWM error closest to the median of its type among clips with above-median displacement.}
\label{fig:real_qual}
\vspace{-3mm}
\end{figure}

LaWM lowers the mean error of the strongest baseline, the Transformer, from $0.615$ to $0.368$, a reduction of $40\%$, and the median from $0.514$ to $0.266$ (Table~\ref{tab:main} and Fig.~\ref{fig:real_qual}).
The improvement is significant against all eight baselines under the paired bootstrap, and against seven at the seed level, the exception being the Neural ODE, one of whose seeds diverges numerically.
DINO-WM, which predicts frozen DINOv2 features, reaches $0.681$; tracking on the real future features at its token resolution already gives $0.589$, so part of its error stems from the feature resolution (\supp{app:dinowm}).
LaWM is the most accurate system on ten of the twelve motion types, with FVIN best on deformation and DINO-WM on deceleration, and its divergence rate is $5.7\%$ against $16.2\%$ to $24.0\%$ for the baselines.
The largest gains occur on uniform, circular, and size-changing motion, where the global readout localizes the object poorly.

\subsection{Generalization and Long-Horizon Prediction}
\label{sec:exp_general}

\begin{table}[!t]
\centering
\small
\setlength{\tabcolsep}{2.5pt}
\caption{Generalization, robustness, and long-horizon prediction of complete systems. The baseline columns report the best conventional system in each setting (the Transformer, except FVIN on pendulum length and the LNN on measured labels). Normalized $xy$ error with mean $\pm$ standard deviation over $5$ seeds, median, and divergence rate (\%). The settings are described in Sec.~\ref{sec:exp_general}.}
\label{tab:general}
\resizebox{\ifdim\width>\columnwidth\columnwidth\else\width\fi}{!}{%
\begin{tabular}{lcccccc}
\toprule
& \multicolumn{3}{c}{Best baseline} & \multicolumn{3}{c}{LaWM (ours)} \\
\cmidrule(lr){2-4}\cmidrule(lr){5-7}
Setting & Mean & Med. & Div. & Mean & Med. & Div. \\
\midrule
Standard & 0.615$\pm$0.034 & 0.514 & 16.2 & \bfseries \best{0.368$\pm$0.012} & \bfseries \best{0.266} & \bfseries \best{5.7} \\
\midrule
\multicolumn{7}{l}{\emph{Data efficiency}} \\
10\% data & 0.658$\pm$0.015 & 0.549 & 20.3 & \bfseries \best{0.504$\pm$0.061} & \bfseries \best{0.367} & \bfseries \best{11.2} \\
25\% data & 0.627$\pm$0.009 & 0.530 & 17.7 & \bfseries \best{0.426$\pm$0.014} & \bfseries \best{0.327} & \bfseries \best{6.0} \\
\midrule
\multicolumn{7}{l}{\emph{Extrapolation}} \\
Time ($5\!\to\!10$) & 0.635$\pm$0.014 & 0.549 & 16.7 & \bfseries \best{0.431$\pm$0.007} & \bfseries \best{0.337} & \bfseries \best{6.8} \\
Unseen types & 0.773$\pm$0.052 & 0.714 & 22.9 & \bfseries \best{0.436$\pm$0.020} & \bfseries \best{0.357} & \bfseries \best{6.0} \\
Pend.\ length & 0.839$\pm$0.019 & 0.825 & 37.1 & \bfseries \best{0.386$\pm$0.044} & \bfseries \best{0.351} & \bfseries \best{1.7} \\
Launch speed & 0.758$\pm$0.094 & 0.722 & 23.1 & \bfseries \best{0.438$\pm$0.026} & \bfseries \best{0.317} & \bfseries \best{10.9} \\
All quantities & 0.762$\pm$0.033 & 0.689 & 25.5 & \bfseries \best{0.465$\pm$0.040} & \bfseries \best{0.356} & \bfseries \best{10.1} \\
\midrule
\multicolumn{7}{l}{\emph{Robustness}} \\
Measured labels & 0.769$\pm$0.009 & 0.776 & 21.1 & \bfseries \best{0.453$\pm$0.018} & \bfseries \best{0.320} & \bfseries \best{9.2} \\
Dilated prompt & 0.615$\pm$0.035 & 0.512 & 16.2 & \bfseries \best{0.394$\pm$0.012} & \bfseries \best{0.292} & \bfseries \best{5.8} \\
\midrule
\multicolumn{7}{l}{\emph{Long horizon}} \\
42 frames & 0.328$\pm$0.065 & 0.267 & \best{2.5} & \bfseries \best{0.247$\pm$0.038} & \bfseries \best{0.155} & \bfseries 3.3 \\
\bottomrule
\end{tabular}}
\vspace{-3mm}
\end{table}

Table~\ref{tab:general} repeats the comparison under harder conditions, with per-method results in \supp{app:general_full}.
In every setting, LaWM is significantly more accurate than all seven conventional baselines under the paired bootstrap.
With $25\%$ of the training clips, LaWM is more accurate than every baseline trained on the full set.
When uniform, decelerated, and circular motion are held out of training, its error rises only from $0.368$ to $0.436$, while the best baseline rises to $0.773$, and the MLP and the Neural ODE diverge on most unseen clips.
When training covers one range of a measured physical quantity and testing lies outside it, LaWM reaches $0.386$ against $0.839$ for pendulum length and $0.438$ against $0.758$ for launch speed, and $0.465$ against $0.762$ when every motion type is extrapolated at once (\supp{app:phys_extrap}).
The comparison is unchanged on annotated positions and with a coarse first-frame prompt (\supp{app:robustness}).
For $42$-frame prediction of damped oscillation, a horizon curriculum selected on validation data is applied to LaWM and, whenever it lowers their validation error, to the baselines (\supp{app:long_horizon}).
LaWM then reaches $0.247$ against $0.328$ for the best baseline.

\subsection{Perception, Transition, and Ablations}
\label{sec:exp_decomp}

\begin{table}[!t]
\centering
\small
\setlength{\tabcolsep}{3pt}
\caption{Separating perception from transition. \textbf{(a)} The transition is fixed to LaWM and the perception varies, with Loc.\ the localization error in pixels. \textbf{(b)} All transitions receive our tracked coordinates and are grouped into black-box, continuous-time, and physics-structured transitions. Normalized $xy$ error over $5$ seeds.}
\label{tab:same_transition}
\phantomsection\label{tab:same_perception}
\resizebox{\ifdim\width>\columnwidth\columnwidth\else\width\fi}{!}{%
\begin{tabular}{lcccc}
\toprule
\multicolumn{5}{l}{\emph{(a) Same transition (LaWM), different perception}} \\
Perception & Loc. & Mean & Median & Div. \\
\midrule
Global features & 89.5 & 0.730$\pm$0.004 & 0.580 & 24.2 \\
\quad + first-frame corr. & \secondbest{61.0} & 0.628$\pm$0.012 & 0.530 & \secondbest{16.8} \\
Learned head & 72.0 & \secondbest{0.603$\pm$0.005} & \secondbest{0.462} & 18.2 \\
Prompted tracking (ours) & \bfseries \best{19.4} & \bfseries \best{0.368$\pm$0.012} & \bfseries \best{0.266} & \bfseries \best{5.7} \\
\midrule
\multicolumn{5}{l}{\emph{(b) Same perception (prompted tracking), different transition}} \\
Transition & & Mean & Median & \\
\midrule
MLP & & \best{0.355$\pm$0.005} & \secondbest{0.280} & \\
GRU & & 0.362$\pm$0.004 & 0.284 & \\
Transformer & & \secondbest{0.361$\pm$0.014} & 0.281 & \\
\addlinespace[2pt]
Neural ODE & & \secondbest{0.361$\pm$0.010} & 0.287 & \\
\addlinespace[2pt]
LNN & & 0.398$\pm$0.006 & 0.326 & \\
VIN & & 0.420$\pm$0.011 & 0.338 & \\
FVIN & & 0.377$\pm$0.007 & 0.296 & \\
LaWM (ours) & & \bfseries 0.368$\pm$0.012 & \bfseries \best{0.266} & \\
\bottomrule
\end{tabular}}
\vspace{-3mm}
\end{table}

Table~\ref{tab:same_transition}(a) fixes the transition and varies the perception.
Prompted tracking reduces the localization error by a factor of three and the prediction error by $39\%$ relative to the best alternative.
Table~\ref{tab:same_transition}(b) fixes the perception.
Given the same coordinates, all transitions except LNN and VIN lie within $0.022$ of each other.
LaWM attains the lowest median and is significantly more accurate than VIN and LNN, while the second-order MLP has a lower mean by $0.013$, which the paired bootstrap finds significant and the seed-level test does not.
With the same coordinates over $42$ frames, the Transformer and a first-order MLP are more accurate than LaWM under direct training, whereas LaWM attains the lowest error ($0.247$) under the validation-selected horizon curriculum (\supp{app:general_full}).
The advantage of the complete system therefore stems mainly from perception, and the forced variational integrator matches black-box accuracy while keeping an explicit Lagrangian and forces.
Under physical extrapolation with the same coordinates, LaWM is the most accurate on four of the five splits (including slope and all-type extrapolation), trailing only FVIN on deceleration rate (\supp{app:phys_extrap}).
On motion types held out during training, the physics-structured transitions attain lower mean errors ($0.375$ to $0.436$) than the black-box transitions and the Neural ODE ($0.439$ to $0.488$), with VIN the lowest; the paired bootstrap confirms the difference for LaWM against the Transformer and the Neural ODE (\supp{app:general_full}).

\paragraph{Inferred physics.}
The learned Lagrangian and forces are explicit and relate to measured physical quantities, although they are never fitted to them.
On the $47$ ramp clips with measured angle, the learned conservative acceleration correlates with the sine of the slope angle (Spearman $\rho=0.72$, $95\%$ CI $[0.56, 0.84]$), and on the $48$ pendulum clips the learned squared frequency correlates with $g/L$ ($\rho=0.54$, $95\%$ CI $[0.31, 0.72]$), although absolute scales are not recovered.
Dissipative motion receives larger viscous coefficients than uniform motion in all five seeds, with median magnitudes of $0.68\pm0.09$ for damped oscillation and $0.63\pm0.10$ for decelerated motion against $0.14\pm0.04$ for uniform motion (\supp{app:param_recovery}).

\paragraph{Ablations.}
Removing both the forces and the context inference, which reduces LaWM to a VIN, significantly raises the error from $0.368$ to $0.420$, while removing only one of them leaves the mean within the seed variance (\supp[Table]{tab:ablation}).
Adding learned forces to the conservative integrator (VIN to FVIN) significantly lowers the error, most on damped oscillation, deformation, and rotation.
A LaWM rollout is slower than a black-box rollout but costs $0.21$\,ms per clip in batch mode (\supp{app:timing}).

\subsection{Action-Conditioned Prediction and Planning}
\label{sec:exp_robot}

\begin{figure}[t]
\centering
\includegraphics[width=\columnwidth]{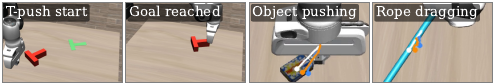}
\caption{LaWM with robot actions in simulation. Left: T-pushing with LaWM in a model-predictive controller. Right: predicted object trajectories (orange) against ground truth (white) and a black-box baseline (blue).}
\label{fig:robot_main}
\vspace{-3mm}
\end{figure}

Because external forces enter the update as an explicit channel, LaWM accepts robot actions without architectural change.
In simulated Franka Panda scenes with randomized mass and friction, it predicts the action-conditioned motion of rigid and deformable objects and, inside a model-predictive controller, pushes a T-shaped block to a goal pose with $34.2\%/39.2\%$ (in-distribution/unseen) success (Fig.~\ref{fig:robot_main}, \supp{app:robot}).

\section{Conclusion}
\label{sec:conclusion}

We presented LaWM, a visual world model whose transition is the forced discrete Euler--Lagrange equation of a Lagrangian and of nonconservative forces inferred from the observed context.
LaWM reads generalized coordinates from real video with a frozen foundation model and prompted tracking, and a single model covers twelve motion types without motion labels.
As a complete system, it reduces the trajectory error and the divergence rate of conventional perception and prediction systems, and it retains this advantage with less training data, beyond the training horizon, on unseen motion types, outside the training range of measured physical quantities, and over $42$-frame rollouts.
Given the same coordinates, it matches black-box transitions while keeping an explicit Lagrangian, explicit forces, and an explicit channel through which actions enter the model.
These components carry physical meaning: adding learned forces to a conservative integrator most improves dissipative motion, the learned potential correlates with measured slope angles, and with the same coordinates the physics-structured transitions transfer best to unseen motion types.

\paragraph{Limitations.}
Because each step solves an implicit equation, a rollout is slower than with a black-box transition.
The model describes one object by a low-dimensional state and does not represent interactions between several objects, and extending it to multi-object scenes is a direction for future work.

\newpage

\clearpage
{
    \small
    \bibliographystyle{ieeenat_fullname}
    \bibliography{references}
}

\clearpage
\setcounter{page}{1}
\onecolumn
\begin{center}
{\Large\textbf{\thetitle}\\[0.5em]Supplementary Material}
\end{center}
\vspace{0.5em}

\appendix
\raggedbottom
\raggedcolumns
\providecommand{\tbd}{\textcolor{red}{TBD}}
\providecommand{\figplaceholder}[2]{\fbox{\parbox[c][#2][c]{0.95\linewidth}{\centering\small\textcolor{red}{#1}}}}

\begin{table}[!b]
\centering
\small
\setlength{\tabcolsep}{3.5pt}
\caption{Per-motion results with the same tracked coordinates and different transitions (full version of Table~3(b) of the main paper).}
\label{tab:same_perception_full}
\resizebox{\ifdim\width>\textwidth\textwidth\else\width\fi}{!}{%
\begin{tabular}{lccccccccccc}
\toprule
& \multicolumn{5}{c}{Black-box} & Continuous & \multicolumn{4}{c}{Physics-structured} & \\
\cmidrule(lr){2-6}\cmidrule(lr){7-7}\cmidrule(lr){8-11}
Motion & MLP & MLP (1st) & GRU & Transformer & RSSM & Neural ODE & HNN & LNN & VIN & FVIN & LaWM (ours) \\
\midrule
Uniform & \secondbest{0.347} & 0.353 & 0.368 & 0.377 & 0.370 & 0.353 & 0.382 & 0.363 & 0.384 & \best{0.341} & \bfseries 0.383 \\
Acceleration & 0.312 & 0.314 & 0.316 & 0.324 & 0.330 & 0.315 & \best{0.302} & 0.404 & 0.381 & 0.334 & \bfseries \secondbest{0.308} \\
Deceleration & 0.304 & 0.302 & 0.305 & 0.301 & \best{0.278} & 0.303 & 0.375 & 0.302 & 0.306 & 0.299 & \bfseries \secondbest{0.296} \\
Parabolic & 0.384 & 0.381 & 0.379 & 0.378 & 0.400 & 0.382 & 0.439 & 0.380 & 0.391 & \secondbest{0.376} & \bfseries \best{0.374} \\
Parabolic + rotation & 0.334 & 0.332 & 0.336 & 0.368 & \best{0.314} & 0.335 & 0.364 & 0.392 & 0.351 & 0.331 & \bfseries \secondbest{0.328} \\
3D & 0.238 & 0.224 & \secondbest{0.218} & 0.232 & 0.220 & 0.248 & 0.266 & 0.261 & 0.267 & 0.240 & \bfseries \best{0.214} \\
Size change & \best{0.527} & 0.562 & 0.554 & 0.552 & 0.574 & 0.551 & 0.572 & 0.614 & 0.570 & 0.552 & \bfseries \secondbest{0.549} \\
Slope & 0.264 & 0.268 & 0.263 & \secondbest{0.262} & 0.267 & 0.270 & 0.274 & 0.360 & 0.312 & 0.279 & \bfseries \best{0.259} \\
Circular & \secondbest{0.363} & 0.379 & 0.386 & 0.375 & 0.390 & \best{0.361} & 0.417 & 0.421 & 0.420 & 0.441 & \bfseries 0.389 \\
Rotation & 0.352 & 0.350 & 0.348 & \secondbest{0.347} & 0.359 & 0.366 & 0.379 & 0.350 & 0.413 & 0.357 & \bfseries \best{0.344} \\
Damped oscillation & 0.327 & 0.369 & 0.332 & \secondbest{0.323} & 0.361 & \best{0.322} & 0.356 & 0.361 & 0.491 & 0.391 & \bfseries 0.375 \\
Deformation & 0.268 & 0.244 & 0.248 & 0.242 & 0.287 & 0.276 & 0.317 & 0.276 & 0.338 & \secondbest{0.239} & \bfseries \best{0.236} \\
\midrule
Mean & \best{0.355$\pm$0.005} & 0.368$\pm$0.005 & 0.362$\pm$0.004 & \secondbest{0.361$\pm$0.014} & 0.376$\pm$0.009 & \secondbest{0.361$\pm$0.010} & 0.388$\pm$0.005 & 0.398$\pm$0.006 & 0.420$\pm$0.011 & 0.377$\pm$0.007 & \bfseries 0.368$\pm$0.012 \\
Median & \secondbest{0.280} & 0.290 & 0.284 & 0.281 & 0.285 & 0.287 & 0.295 & 0.326 & 0.338 & 0.296 & \bfseries \best{0.266} \\
Div.\ (\%) & 5.8 & 6.3 & 6.0 & 6.4 & \best{4.5} & 5.6 & \secondbest{5.1} & 6.1 & 7.2 & 6.2 & \bfseries 5.7 \\
\bottomrule
\end{tabular}}
\end{table}

\begin{multicols}{2}

\section*{Overview}
\label{app:overview}
The supplementary material is organized as follows. Appendix~\ref{app:per_motion} compares all transitions on the same tracked coordinates, with the additional and classical baselines and the per-type statistics (Sec.~4.1 and~4.4). Appendix~\ref{app:long_horizon_all} reports long-horizon consistency and the horizon curriculum (Sec.~4.3). Appendix~\ref{app:generalization} gives the generalization results per method, the extrapolation in measured physical quantities, and the robustness to the label source and to coarse prompts (Sec.~4.3). Appendix~\ref{app:analysis} analyzes the learned physics through ablations, measured quantities and dissipation, force fields, the physics code, and hyperparameter sensitivity (Sec.~4.4). Appendix~\ref{app:perception} covers perception, including the DINO-WM baseline (Sec.~4.2), off-the-shelf trackers, and video generation models. Appendix~\ref{app:robot} describes the simulated manipulation experiments (Sec.~4.5). Appendix~\ref{app:details} gives the dataset annotations, implementation details, the statistical protocol (Sec.~4.1), and the inference cost.
In every table, dark and light blue mark the best and second-best results within the row, column, or block that the table compares, a single blue cell marks the better of two compared results, and our results are in bold. Unless stated otherwise, errors are normalized $xy$ errors.

LaWM is an object-centric physical world model. From real video, it predicts the generalized coordinates of a prompted object by integrating a Lagrangian and nonconservative forces inferred from the observed context. It is compared with black-box, continuous-time, and physics-structured transitions, with a feature-space world model (DINO-WM), and with video generation models on the same real test split and with the same metrics. Its explicit force channel $B(\eta)u_k$ connects it directly to model-predictive control (Appendix~\ref{app:robot}).

\section{Same-Coordinate Comparison of Transitions}
\label{app:per_motion}

This appendix isolates the transition rule by giving every method the same tracked coordinates. In the standard setting, all transitions other than HNN, LNN, and VIN lie within $0.022$ of one another in mean error. LaWM attains the lowest median ($0.266$, Table~\ref{tab:same_perception_full}) and is significantly more accurate than HNN, LNN, and VIN (Table~\ref{tab:overall_sig}).
The transitions separate as the setting becomes harder. LaWM attains the lowest mean error with $10\%$ and $25\%$ of the training data ($0.504$ and $0.426$) and under temporal extrapolation ($0.431$, Table~\ref{tab:general_full}). It is also the most accurate in four of the five extrapolation splits of measured physical quantities and lies below every black-box and continuous-time transition in all five (Table~\ref{tab:phys_extrap}).
On motion types held out during training, the physics-structured transitions form the most accurate group, with errors from $0.375$ to $0.436$, ahead of every black-box transition and the Neural ODE, which range from $0.439$ to $0.488$ (Table~\ref{tab:general_full}).
Over long horizons on real IRIS windows, LaWM attains the lowest error from frame $11$ and the lowest divergence rate from frame $18$ up to frame $64$, and at frame $32$ it is significantly more accurate than every other transition (Fig.~\ref{fig:horizon_curve}). It also attains the lowest error for $42$-frame prediction under the validation-selected protocol ($0.247$, Appendix~\ref{app:long_horizon}), and in $94\%$ of free rollouts on passive clips its energy never exceeds the initial value (Fig.~\ref{fig:free_rollout}).
The physical structure thus helps most where a world model is actually used, namely with limited data, outside the training range, and over long rollouts.

\subsection{Per-Motion Results}
\label{app:per_motion_results}
\label{app:extra_baselines}

Table~\ref{tab:same_perception_full} gives the per-motion results. Every method receives the coordinates from our prompted tracker, so only the transition rule differs. Metrics and seeds follow Table~1 of the main paper. As in Table~1 of the main paper, the Mean, Median, and Div.\ rows leave out the Parabolic + rotation row ($7$ Falling Objects clips) and are computed over the other $554$ test clips.
LaWM has the lowest error on five of the twelve motion types (parabolic, 3D, slope, rotation, and deformation) and the second lowest on four more. Its largest gaps are on uniform motion, circular motion, and damped oscillation, where FVIN or the Neural ODE is ahead by $0.03$ to $0.05$. Over all clips the MLP has the lowest mean ($0.355$) and LaWM the lowest median ($0.266$), so LaWM is more accurate on the typical clip but has a heavier tail of large errors.

Table~\ref{tab:extra_baselines} adds HNN and RSSM, evaluated on the standard test set over $5$ seeds. In the complete-system columns they use the conventional perception of Table~1 of the main paper, and with it both stay above $0.66$ in mean error, against $0.368$ for LaWM. With tracked coordinates RSSM reaches $0.376$ and the lowest divergence rate ($4.5\%$), and HNN reaches $0.388$. Most of the gap between these complete systems and LaWM therefore comes from perception, and with the same coordinates the three transitions lie within $0.020$ of one another.

\begin{table}[H]
\centering
\small
\setlength{\tabcolsep}{3pt}
\caption{Additional baselines on the standard test set.}
\label{tab:extra_baselines}
\resizebox{\ifdim\width>\columnwidth\columnwidth\else\width\fi}{!}{%
\begin{tabular}{lcccccc}
\toprule
& \multicolumn{3}{c}{Complete system} & \multicolumn{3}{c}{Tracked coordinates} \\
\cmidrule(lr){2-4}\cmidrule(lr){5-7}
Transition & Mean & Med. & Div. & Mean & Med. & Div. \\
\midrule
HNN & 0.680$\pm$0.012 & 0.591 & 18.6 & 0.388$\pm$0.005 & 0.295 & \secondbest{5.1} \\
RSSM & \secondbest{0.661$\pm$0.016} & \secondbest{0.561} & \secondbest{17.9} & \secondbest{0.376$\pm$0.009} & \secondbest{0.285} & \best{4.5} \\
LaWM (ours) & \bfseries \best{0.368$\pm$0.012} & \bfseries \best{0.266} & \bfseries \best{5.7} & \bfseries \best{0.368$\pm$0.012} & \bfseries \best{0.266} & \bfseries 5.7 \\
\bottomrule
\end{tabular}}
\end{table}

\subsection{Classical Baselines}
\label{app:classical}

Table~\ref{tab:classical} compares LaWM with classical predictors that receive the same tracked coordinates and context as the learned transitions. The error is taken over $10$ predicted frames on the $554$ test clips. The analytic fit selects the closed-form model of the ground-truth motion type, so it uses the label and serves only as a reference.
Constant velocity and the constant-velocity Kalman filter are the strongest classical predictors, at about $0.45$ mean error. Both constant-acceleration models do worse than simply keeping the last position, which suggests that an acceleration estimated from six noisy frames amplifies the tracking noise. Even with the motion label, the analytic fit reaches only $0.550$, while LaWM reaches $0.368$ without any label. The Kalman filter changes the constant-velocity prediction only slightly ($0.446$ against $0.449$), and every classical predictor diverges on at least $8.8\%$ of the clips, against $5.7\%$ for LaWM.

\begin{table}[H]
\centering
\small
\setlength{\tabcolsep}{3pt}
\caption{Classical predictors on real video with the same tracked coordinates.}
\label{tab:classical}
\resizebox{\ifdim\width>\columnwidth\columnwidth\else\width\fi}{!}{%
\begin{tabular}{lcccc}
\toprule
Method & Motion label & Mean & Median & Div.\ (\%) \\
\midrule
Static (last position) & No & 0.616 & 0.553 & 17.1 \\
Constant velocity & No & 0.449 & \secondbest{0.323} & 9.7 \\
Constant acceleration & No & 0.744 & 0.460 & 22.0 \\
Kalman filter (constant velocity) & No & \secondbest{0.446} & 0.331 & \secondbest{8.8} \\
Kalman filter (constant acceleration) & No & 0.720 & 0.449 & 20.2 \\
\midrule
Analytic fit per motion type & Yes & 0.550 & 0.404 & 14.3 \\
\midrule
LaWM (ours) & No & \bfseries \best{0.368$\pm$0.012} & \bfseries \best{0.266} & \bfseries \best{5.7} \\
\bottomrule
\end{tabular}}
\end{table}

\subsection{Orientation, Size, and Shape}
\label{app:full_state}

Table~\ref{tab:full_state} reports the errors on the remaining generalized coordinates, with the mean over $5$ seeds. Each column averages over the motion types in which that coordinate changes, namely orientation (parabolic + rotation, rotation), size (3D, size change), and shape (deformation).
Across these coordinates, LaWM consistently ranks among the top two transitions: it achieves the lowest error on size ($34.8\%$, ahead of Transformer at $35.2\%$) and the second-lowest error on both orientation ($39.5^\circ$, behind HNN at $39.2^\circ$) and shape ($18.1\%$, behind FVIN at $17.3\%$). Unlike other baselines that excel on a single coordinate while degrading on others, LaWM maintains strong performance across all generalized coordinates in addition to object position.

\begin{table}[H]
\centering
\small
\setlength{\tabcolsep}{4pt}
\caption{Errors on the remaining generalized coordinates with the same tracked coordinates.}
\label{tab:full_state}
\begin{tabular}{lccc}
\toprule
Transition & Orientation ($^\circ$) & Size (\%) & Shape (\%) \\
\midrule
MLP & 40.2 & 35.8 & 20.2 \\
MLP (1st) & 40.7 & 35.6 & 20.4 \\
GRU & 41.1 & 36.4 & 19.7 \\
Transformer & 40.1 & \secondbest{35.2} & 18.5 \\
RSSM & 40.3 & 37.3 & 18.8 \\
Neural ODE & 39.8 & 37.0 & 20.0 \\
HNN & \best{39.2} & 38.8 & 20.1 \\
LNN & 41.6 & 37.4 & 19.9 \\
VIN & 41.9 & 41.3 & 23.1 \\
FVIN & 44.8 & 39.3 & \best{17.3} \\
LaWM (ours) & \bfseries \secondbest{39.5} & \bfseries \best{34.8} & \bfseries \secondbest{18.1} \\
\bottomrule
\end{tabular}
\end{table}

\subsection{Per-Type and Overall Statistics}
\label{app:per_type}

Table~\ref{tab:per_type} compares, for each motion type, the error of LaWM with that of the best other transition in Table~\ref{tab:same_perception_full}. It gives $95\%$ bootstrap confidence intervals over test clips and Holm-corrected paired-bootstrap $p$-values. After the correction, only damped oscillation shows a significant difference ($p=0.008$, in favor of the Neural ODE). In every other row the confidence intervals overlap and $p\ge0.096$, so most per-type rankings in Table~\ref{tab:same_perception_full} lie within the noise across clips.
Table~\ref{tab:overall_sig} reports the overall paired bootstrap on the $554$ test clips over $5$ seeds. $\Delta = \text{LaWM} - \text{Baseline}$ is the difference in mean error, so a negative value favors LaWM. LaWM is significantly more accurate than HNN, LNN, and VIN, significantly less accurate than the MLP ($+0.013$, $p=0.004$), and not significantly different from the remaining six transitions.

\begin{table}[H]
\centering
\small
\setlength{\tabcolsep}{2.5pt}
\captionof{table}{Per-motion significance with the same tracked coordinates.}
\label{tab:per_type}
\resizebox{\ifdim\width>\columnwidth\columnwidth\else\width\fi}{!}{%
\begin{tabular}{lrclcr}
\toprule
Motion & Clips & LaWM [95\% CI] & Best other & Best other [95\% CI] & $p$ (Holm) \\
\midrule
Uniform & 64 & \bfseries 0.383 [0.326, 0.441] & FVIN & \best{0.341 [0.309, 0.377]} & 0.096 \\
Acceleration & 50 & \bfseries 0.308 [0.256, 0.366] & HNN & \best{0.302 [0.254, 0.354]} & 1.000 \\
Deceleration & 18 & \bfseries 0.296 [0.241, 0.362] & RSSM & \best{0.278 [0.213, 0.349]} & 0.720 \\
Parabolic & 51 & \bfseries \best{0.374 [0.279, 0.500]} & FVIN & 0.376 [0.270, 0.494] & 1.000 \\
Parabolic + rotation & 7 & \bfseries 0.328 [0.235, 0.428] & RSSM & \best{0.314 [0.260, 0.374]} & 1.000 \\
3D & 25 & \bfseries \best{0.214 [0.172, 0.261]} & GRU & 0.218 [0.175, 0.265] & 1.000 \\
Size change & 83 & \bfseries 0.549 [0.478, 0.623] & MLP & \best{0.527 [0.460, 0.602]} & 0.248 \\
Slope & 50 & \bfseries \best{0.259 [0.214, 0.312]} & Transformer & 0.262 [0.221, 0.307] & 1.000 \\
Circular & 42 & \bfseries 0.389 [0.304, 0.488] & Neural ODE & \best{0.361 [0.291, 0.439]} & 0.284 \\
Rotation & 50 & \bfseries \best{0.344 [0.286, 0.409]} & Transformer & 0.347 [0.290, 0.410] & 1.000 \\
Damped oscillation & 97 & \bfseries 0.375 [0.319, 0.436] & Neural ODE & \best{0.322 [0.286, 0.362]} & 0.008 \\
Deformation & 24 & \bfseries \best{0.236 [0.181, 0.295]} & FVIN & 0.239 [0.171, 0.315] & 1.000 \\
\bottomrule
\end{tabular}}
\end{table}

\begingroup
\addtocounter{table}{-1}
\renewcommand{\thetable}{\arabic{table}b}
\makeatletter
\@ifundefined{theHtable}{}{\renewcommand{\theHtable}{\arabic{table}b}}
\makeatother
\begin{table}[H]
\centering
\small
\setlength{\tabcolsep}{3pt}
\caption{Overall paired-bootstrap comparison with the same tracked coordinates.}
\label{tab:overall_sig}
\resizebox{\ifdim\width>\columnwidth\columnwidth\else\width\fi}{!}{%
\begin{tabular}{lccc}
\toprule
Baseline & $\Delta$ (LaWM $-$ Baseline) & 95\% CI & $p$-value \\
\midrule
MLP & $+0.013$ & $[+0.004,\ +0.022]$ & $0.004$ \\
MLP (1st) & $+0.000$ & $[-0.011,\ +0.011]$ & $0.962$ \\
GRU & $+0.006$ & $[-0.004,\ +0.016]$ & $0.241$ \\
Transformer & $+0.007$ & $[-0.005,\ +0.019]$ & $0.268$ \\
RSSM & $-0.008$ & $[-0.020,\ +0.004]$ & $0.194$ \\
Neural ODE & $+0.007$ & $[-0.004,\ +0.018]$ & $0.218$ \\
HNN & $-0.020$ & $[-0.032,\ -0.008]$ & $0.001$ \\
LNN & $-0.030$ & $[-0.043,\ -0.017]$ & $<0.001$ \\
VIN & $-0.052$ & $[-0.068,\ -0.036]$ & $<0.001$ \\
FVIN & $-0.009$ & $[-0.021,\ +0.003]$ & $0.142$ \\
\bottomrule
\end{tabular}}
\end{table}
\endgroup

\section{Long-Horizon Consistency}
\label{app:long_horizon_all}

\subsection{Error versus Prediction Horizon}
\label{app:horizon_curve}

\begin{figure}[H]
\centering
\includegraphics[width=0.88\columnwidth]{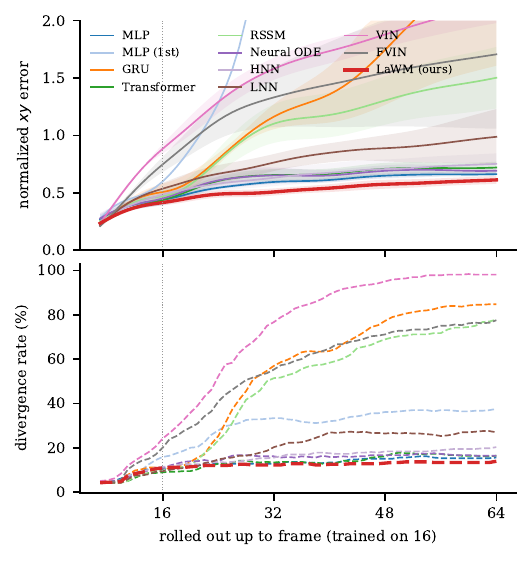}
\caption{Error (top) and divergence rate (bottom) versus prediction horizon on $72$ real $128$-frame IRIS windows.}
\label{fig:horizon_curve}
\end{figure}

Figure~\ref{fig:horizon_curve} extends the prediction beyond the training horizon on real clips with ground truth. Every model is trained on $16$-frame clips, observes the first $6$ frames, and then predicts autoregressively up to frame $64$ without any correction, with the same tracked coordinates for every transition. The error at a given frame is taken over all predicted frames up to that frame and averaged over clips and $5$ seeds, and the shaded band is the standard deviation over seeds.
LaWM attains the lowest error at every horizon from frame $11$ and the lowest divergence rate at every horizon from frame $18$ up to frame $64$. Its error is $0.507$ at frame $32$ and $0.612$ at frame $64$, against $0.594$ and $0.662$ for the MLP, $0.655$ and $0.720$ for the Transformer, and $0.763$ and $0.987$ for the LNN.
The paired bootstrap over clips finds LaWM significantly more accurate than every other transition at frame $32$, and than every other transition except the MLP at frame $64$.
VIN and FVIN, which share the integrator but not the inferred physics code, diverge on more than half of the clips from frame $32$ on.

\subsection{Free Rollouts and Energy}
\label{app:free_rollout}

\begin{figure*}[!t]
\centering
\includegraphics[width=\textwidth]{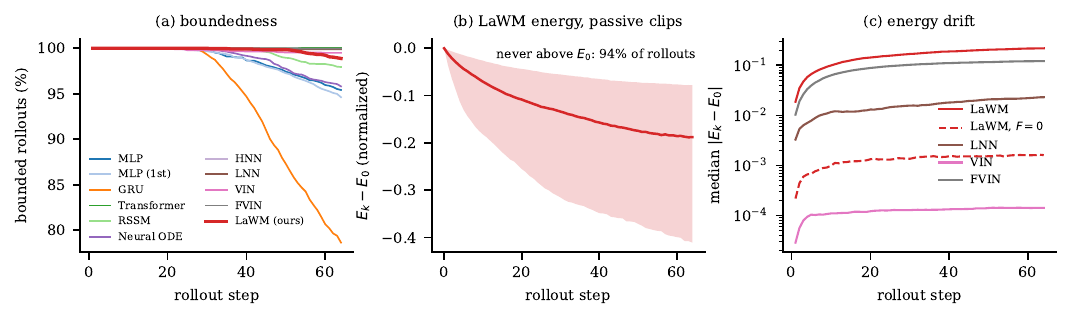}
\caption{Free rollouts over $64$ steps on the $554$ test clips. (a) Fraction of bounded rollouts. (b) Energy of LaWM rollouts on passive clips. (c) Median energy drift.}
\label{fig:free_rollout}
\end{figure*}

Figure~\ref{fig:free_rollout} rolls out every transition for $64$ steps without ground truth, which matches the prediction range of Fig.~\ref{fig:horizon_curve}. Panel (a) shows the fraction of rollouts that remain bounded at each step. A rollout that stops moving also counts as bounded, so this panel measures stability rather than accuracy. Panel (b) shows the energy of LaWM rollouts on passive clips relative to the last context step, and in $94\%$ of these rollouts the energy never exceeds its initial value. Panel (c) shows the median energy drift for LaWM with and without the learned forces and for the LNN, VIN, and FVIN. With the learned forces, the energy of LaWM decreases through dissipation, as expected for passive motion with friction.

\subsection{Long-Horizon Training Protocol}
\label{app:long_horizon}

For $42$-frame damped oscillation prediction, training directly on the full rollout horizon can destabilize early optimization. As noted in Sec.~4.4 of the main paper, under direct training without a curriculum (\emph{no curr.} in Table~\ref{tab:general_full}), the first-order MLP ($0.264$) and Transformer ($0.284$) reach a lower error than LaWM ($0.298$) with the same tracked coordinates. We therefore evaluate a horizon curriculum that lengthens the supervised rollout during training, with the schedule selected on the validation split. For a fair comparison, the same curriculum is also evaluated for every baseline. Table~\ref{tab:general_full} reports direct training without the curriculum and, for each method, the variant selected on validation (with or without the curriculum). Under this protocol LaWM attains the lowest error ($0.247$).

\subsection{Kinematic Consistency}
\label{app:phys_consistency}

Table~\ref{tab:phys_consistency} reports the error of the fitted velocity (uniform motion) or acceleration (other columns) against the ground truth, in normalized units per frame for the velocity and per frame$^2$ for the acceleration, both in units of $10^{-2}$ and averaged over clips and $5$ seeds. This evaluation checks whether predicted trajectories preserve physically accurate derivatives, which a low position error alone does not guarantee. The top block compares complete systems, and in the bottom block every transition receives the tracked coordinates.

Among complete systems, LaWM attains the lowest error across all five motion types: uniform ($7.5$, ahead of FVIN at $9.4$), accelerated ($1.3$, compared to $2.3$ to $4.8$ for the baselines), decelerated ($0.7$, tied with FVIN), parabolic ($1.0$), and slope ($1.5$) motion, while the Neural ODE diverges on uniform motion. With tracked coordinates every baseline improves substantially, for example the MLP drops from $22.3$ to $8.6$ on uniform motion and from $4.3$ to $1.4$ on accelerated motion, while the LaWM row remains identical in both blocks. Even under identical perception, LaWM consistently ranks among the top two transitions in every column: it achieves the lowest error on slope motion ($1.5$, ahead of the Transformer at $1.6$) and ties for best on accelerated motion ($1.3$, with the first-order MLP), while ranking second on uniform ($7.5$, behind the first-order MLP at $7.1$), decelerated ($0.7$, behind FVIN at $0.5$), and parabolic ($1.0$, behind FVIN at $0.9$) motion.

\begin{table}[H]
\centering
\small
\setlength{\tabcolsep}{3pt}
\caption{Kinematic consistency on real test clips ($\times10^{-2}$).}
\label{tab:phys_consistency}
\resizebox{\ifdim\width>\columnwidth\columnwidth\else\width\fi}{!}{%
\begin{tabular}{lccccc}
\toprule
System & Uniform & Accel. & Decel. & Parabolic & Slope \\
\midrule
\multicolumn{6}{l}{\emph{Complete systems}} \\
MLP & 22.3 & 4.3 & 1.9 & 2.7 & 2.7 \\
MLP (1st) & 13.1 & \secondbest{2.3} & 1.3 & \secondbest{1.9} & 2.1 \\
GRU & 13.3 & 3.5 & 2.2 & \secondbest{1.9} & 2.2 \\
Transformer & 10.5 & 2.9 & \secondbest{1.2} & 2.0 & 2.5 \\
Neural ODE & div. & 4.8 & 2.0 & 2.9 & 2.9 \\
LNN & 14.0 & 4.2 & 2.4 & 3.2 & 2.9 \\
VIN & 15.7 & 3.1 & 1.7 & 2.2 & \secondbest{2.0} \\
FVIN & \secondbest{9.4} & 3.1 & \best{0.7} & 2.7 & 2.2 \\
LaWM (ours) & \bfseries \best{7.5} & \bfseries \best{1.3} & \bfseries \best{0.7} & \bfseries \best{1.0} & \bfseries \best{1.5} \\
\midrule
\multicolumn{6}{l}{\emph{Same perception}} \\
MLP & 8.6 & \secondbest{1.4} & 1.2 & 1.1 & 1.8 \\
MLP (1st) & \best{7.1} & \best{1.3} & 0.9 & 1.7 & 1.8 \\
GRU & 8.9 & 2.2 & 1.5 & 1.2 & 1.7 \\
Transformer & 9.7 & 1.5 & 1.4 & 1.1 & \secondbest{1.6} \\
Neural ODE & 8.4 & 1.5 & 1.2 & 1.1 & 1.9 \\
LNN & 8.3 & 2.1 & 1.2 & 1.3 & 1.9 \\
VIN & 8.8 & 2.1 & 0.8 & 1.2 & 2.4 \\
FVIN & 7.8 & 2.4 & \best{0.5} & \best{0.9} & 2.3 \\
LaWM (ours) & \bfseries \secondbest{7.5} & \bfseries \best{1.3} & \bfseries \secondbest{0.7} & \bfseries \secondbest{1.0} & \bfseries \best{1.5} \\
\bottomrule
\end{tabular}}
\end{table}
\subsection{Solver Convergence}
\label{app:solver}

Table~\ref{tab:solver} varies the number of Newton iterations $J$ and reports the error and the mean DEL residual on the test clips over $3$ seeds (seeds $0$ to $2$). The submitted model uses $J=4$. From $J=1$ to $J=8$ the residual falls by a factor of about $45$, from $1.1\times10^{-2}$ to $2.4\times10^{-4}$, while the error stays between $0.362$ and $0.372$, within the seed variance.

\begin{table}[H]
\centering
\small
\setlength{\tabcolsep}{4pt}
\caption{Number of Newton iterations $J$ in the forced discrete Euler--Lagrange solve.}
\label{tab:solver}
\resizebox{\ifdim\width>\columnwidth\columnwidth\else\width\fi}{!}{%
\begin{tabular}{lcccc}
\toprule
Iterations & Mean & Median & Div.\ (\%) & DEL residual \\
\midrule
$J=1$ & 0.369$\pm$0.004 & 0.273 & 5.2 & $1.1\times10^{-2}$ \\
$J=2$ & 0.367$\pm$0.010 & 0.260 & 5.4 & $7.7\times10^{-3}$ \\
$J=4$ (default) & \bfseries 0.362$\pm$0.010 & \bfseries 0.264 & \bfseries 5.3 & \bfseries $1.4\times10^{-3}$ \\
$J=8$ & 0.372$\pm$0.008 & 0.264 & 5.8 & $2.4\times10^{-4}$ \\
\bottomrule
\end{tabular}}
\end{table}

\section{Generalization and Robustness}\label{app:generalization}
\subsection{Generalization Results per Method}\label{app:general_full}

\begin{table*}[!t]
\centering
\small
\setlength{\tabcolsep}{3.5pt}
\caption{Generalization and long-horizon prediction for all methods. Mean normalized $xy$ error, median in parentheses.}
\label{tab:general_full}
\resizebox{\ifdim\width>\textwidth\textwidth\else\width\fi}{!}{%
\begin{tabular}{lcccccc}
\toprule
Method & 10\% data & 25\% data & Extrapolation & Unseen types & Long horizon (no curr.) & Long horizon \\
\midrule
\multicolumn{7}{l}{\emph{Complete systems (conventional perception for baselines)}} \\
MLP & 0.885 (0.702) & 0.783 (0.632) & div.\ (0.603) & div.\ (19.6) & 0.384 (0.318) & 0.368 (0.308) \\
MLP (1st) & 0.774 (0.696) & 0.722 (0.627) & 0.703 (0.587) & div.\ (1.556) & div.\ (0.249) & div.\ (0.237) \\
GRU & 0.954 (0.813) & 0.773 (0.624) & 0.646 (0.560) & 1.407 (1.283) & 0.782 (0.341) & 0.746 (0.325) \\
Transformer & \secondbest{0.658} (0.549) & \secondbest{0.627} (0.530) & \secondbest{0.635} (0.549) & \secondbest{0.773} (0.714) & \secondbest{0.342} (0.278) & \secondbest{0.328} (0.267) \\
Neural ODE & 0.909 (0.723) & 0.779 (0.635) & div.\ (0.617) & div.\ (div.) & 0.402 (0.291) & 0.379 (0.276) \\
LNN & 0.740 (0.661) & 0.701 (0.632) & 0.763 (0.638) & 2.546 (1.433) & 0.614 (0.549) & 0.576 (0.528) \\
VIN & 0.874 (0.746) & 0.881 (0.757) & 0.842 (0.723) & 0.870 (0.700) & 0.735 (0.468) & 0.689 (0.436) \\
FVIN & 0.694 (0.607) & 0.663 (0.594) & 0.675 (0.608) & 1.437 (0.723) & 0.489 (0.342) & 0.437 (0.317) \\
LaWM (ours) & \bfseries \best{0.504} (0.367) & \bfseries \best{0.426} (0.327) & \bfseries \best{0.431} (0.337) & \bfseries \best{0.436} (0.357) & \bfseries \best{0.298} (0.184) & \bfseries \best{0.247} (0.155) \\
\midrule
\multicolumn{7}{l}{\emph{Same perception (tracked coordinates for all)}} \\
MLP & 0.656 (0.391) & 0.436 (0.373) & 0.439 (0.354) & 0.439 (0.389) & 0.346 (0.189) & 0.312 (0.172) \\
MLP (1st) & 0.617 (0.505) & 0.436 (0.359) & \secondbest{0.434} (0.340) & 0.456 (0.408) & \best{0.264} (0.162) & \secondbest{0.253} (0.156) \\
GRU & 0.576 (0.419) & 0.444 (0.363) & 0.446 (0.347) & 0.460 (0.391) & 0.312 (0.204) & 0.285 (0.191) \\
Transformer & 0.515 (0.404) & \secondbest{0.432} (0.331) & 0.436 (0.343) & 0.472 (0.424) & \secondbest{0.284} (0.160) & 0.282 (0.157) \\
Neural ODE & 0.771 (0.388) & 0.443 (0.376) & 0.454 (0.378) & 0.488 (0.424) & 0.378 (0.269) & 0.342 (0.253) \\
LNN & 0.653 (0.428) & 0.460 (0.379) & 0.483 (0.383) & \secondbest{0.432} (0.365) & 0.562 (0.448) & 0.518 (0.412) \\
VIN & \secondbest{0.511} (0.395) & 0.474 (0.375) & 0.469 (0.400) & \best{0.375} (0.314) & 0.648 (0.471) & 0.596 (0.434) \\
FVIN & 0.833 (0.384) & 0.467 (0.342) & 0.438 (0.369) & 0.434 (0.372) & 0.335 (0.214) & 0.287 (0.188) \\
LaWM (ours) & \bfseries \best{0.504} (0.367) & \bfseries \best{0.426} (0.327) & \bfseries \best{0.431} (0.337) & \bfseries 0.436 (0.357) & \bfseries 0.298 (0.184) & \bfseries \best{0.247} (0.155) \\
\bottomrule
\end{tabular}}
\end{table*}

Table~\ref{tab:general_full} reports the mean error with the median in parentheses over $5$ seeds, and ``div.'' marks a mean dominated by numerically diverging seeds. In the top block the baselines use the conventional perception, and in the bottom block every transition receives the tracked coordinates, so the LaWM row is the same in both blocks. For $42$-frame prediction (Appendix~\ref{app:long_horizon}) we report direct training without the horizon curriculum (no curr.) and, for every method, the variant selected on validation.
As a complete system, LaWM has the lowest error in all six settings, while several baselines diverge numerically under extrapolation, on unseen types, or over long horizons. With the same tracked coordinates, LaWM remains the most accurate with $10\%$ and $25\%$ of the data, under temporal extrapolation, and for $42$-frame prediction with the selected protocol. VIN is the most accurate on unseen motion types ($0.375$), and the first-order MLP is the most accurate for $42$-frame prediction without the curriculum ($0.264$).

\subsection{Extrapolation in Measured Physical Quantities}
\label{app:phys_extrap}

Table~\ref{tab:phys_extrap} gives the full five-split evaluation of extrapolation in measured physical quantities under a unified event-level split protocol. Training and validation use one range of the quantity, and testing uses the clips outside that range, split by recording event; other motion types keep their original split. The top block gives every transition the same prompted tracking, and the bottom block compares complete systems, with the global readout refitted on each training split. Errors are averaged over $5$ seeds.
Under this protocol with the same perception, LaWM is the most accurate on four of the five splits (including slope acceleration and all motion types), and FVIN is ahead on the deceleration rate split ($0.290$ against $0.334$), which has only $15$ test clips. On the other four splits the margin over the second-best transition is small, $0.006$ on pendulum length, $0.004$ on launch speed, and $0.005$ on slope acceleration and on all motion types. The gain from the physical structure is therefore consistent but modest when every transition receives the same coordinates. The other transitions are also less even across splits. The LNN reaches $0.695$ on pendulum length and $0.574$ on slope acceleration, and FVIN, the best on deceleration rate, reaches $0.486$ on pendulum length.
As complete systems, the baselines range from $0.469$ to $1.031$, and LaWM is the most accurate on every split, for example $0.386$ against $0.839$ for FVIN on pendulum length and $0.465$ against $0.762$ for the Transformer on all motion types. The baselines lose far more accuracy here than with tracked coordinates, so most of this gap comes from their conventional perception, which is refitted on each training split and transfers poorly outside its range.

\begin{table}[H]
\centering
\small
\setlength{\tabcolsep}{2pt}
\caption{Extrapolation in measured physical quantities across all five splits.}
\label{tab:phys_extrap}
\resizebox{\ifdim\width>\columnwidth\columnwidth\else\width\fi}{!}{%
\begin{tabular}{lcccccccc}
\toprule
Split (test clips) & MLP & GRU & Transformer & Neural ODE & LNN & VIN & FVIN & LaWM \\
\midrule
\multicolumn{9}{l}{\emph{Same perception}} \\
Pendulum length (48) & \secondbest{0.392} & 0.497 & 0.416 & 0.397 & 0.695 & 0.416 & 0.486 & \bfseries \best{0.386} \\
Launch speed (70) & 0.478 & 0.449 & 0.455 & \secondbest{0.442} & 0.513 & 0.489 & 0.477 & \bfseries \best{0.438} \\
Slope acceleration (51) & 0.431 & \secondbest{0.430} & 0.475 & 0.447 & 0.574 & 0.444 & 0.447 & \bfseries \best{0.425} \\
Deceleration rate (15) & 0.385 & 0.339 & 0.409 & 0.348 & 0.394 & 0.366 & \best{0.290} & \bfseries \secondbest{0.334} \\
All motion types (184) & 0.473 & 0.492 & 0.489 & 0.492 & 0.549 & \secondbest{0.470} & 0.497 & \bfseries \best{0.465} \\
\midrule
\multicolumn{9}{l}{\emph{Complete systems}} \\
Pendulum length (48) & 0.899 & 0.888 & 0.864 & 0.912 & 0.960 & 0.958 & \secondbest{0.839} & \bfseries \best{0.386} \\
Launch speed (70) & 0.946 & 0.888 & \secondbest{0.758} & 0.955 & 0.827 & 1.031 & 0.966 & \bfseries \best{0.438} \\
Slope acceleration (51) & 0.702 & \secondbest{0.654} & 0.683 & 0.735 & 0.816 & 0.819 & 0.675 & \bfseries \best{0.425} \\
Deceleration rate (15) & 0.514 & 0.544 & \secondbest{0.469} & 0.520 & 0.473 & 0.640 & 0.567 & \bfseries \best{0.334} \\
All motion types (184) & 0.846 & 0.796 & \secondbest{0.762} & 0.962 & 0.868 & 0.952 & 0.852 & \bfseries \best{0.465} \\
\bottomrule
\end{tabular}}
\end{table}

\subsection{Label Source and Prompt Robustness}
\label{app:robustness}
\label{app:auto_prompt}

Table~\ref{tab:robustness} tests whether the complete-system comparison depends on how the labels and prompts are obtained. The top block keeps only the test clips whose positions come from dataset annotations (MOT17, UA-DETRAC, TbD-3D, Shuttlecock, and GraviCap, $152$ clips). On this subset LaWM reaches $0.453$, against $0.769$ for the best baseline (LNN). The bottom block perturbs the first-frame prompt on all $554$ test clips and evaluates the trained models, and the baselines receive the same first-frame anchor. Loc.\ is the tracking error in pixels. The error of LaWM grows with the tracking error, from $0.346$ with a point prompt to $0.394$ with a mask dilated to twice its area, but stays far below the best baseline ($0.615$ to $0.620$) for every prompt. A point prompt even gives a lower error than the annotated mask ($0.346$ against $0.368$), so LaWM does not need a precise mask to start the prediction.

\begin{table}[H]
\centering
\small
\setlength{\tabcolsep}{3.5pt}
\caption{Robustness of the complete-system comparison to the label source (top) and to the first-frame prompt (bottom).}
\label{tab:robustness}
\resizebox{\columnwidth}{!}{%
\begin{tabular}{lcccc}
\toprule
Setting & Loc.\ (px) & Best baseline & LaWM & LaWM Med. \\
\midrule
Measured-label subset & -- & 0.769 (LNN) & \bfseries \best{0.453$\pm$0.018} & \bfseries 0.320 \\
\midrule
Annotated mask & 19.4 & 0.615 & \bfseries \best{0.368$\pm$0.012} & \bfseries 0.266 \\
Point (disc at centroid) & 14.9 & 0.615 & \bfseries \best{0.346$\pm$0.014} & \bfseries 0.249 \\
Eroded mask (50\% area) & 16.7 & 0.616 & \bfseries \best{0.353$\pm$0.013} & \bfseries 0.256 \\
Bounding box & 22.4 & 0.615 & \bfseries \best{0.374$\pm$0.012} & \bfseries 0.271 \\
Shifted mask & 23.2 & 0.620 & \bfseries \best{0.383$\pm$0.012} & \bfseries 0.289 \\
Dilated mask (200\% area) & 25.6 & 0.615 & \bfseries \best{0.394$\pm$0.012} & \bfseries 0.292 \\
\bottomrule
\end{tabular}}
\end{table}

\section{Analysis of the Learned Physics}
\label{app:analysis}

The learned Lagrangian and forces are explicit, and they organize by physical regime without any supervision on physical quantities.
On ramps, the learned conservative acceleration correlates strongly with the sine of the measured slope angle (Spearman $\rho=0.72$, $95\%$ CI $[0.56, 0.84]$, Table~\ref{tab:param_recovery}), and on Acceleration and Slope clips the learned force field aligns with gravity and with the ramp direction (Fig.~\ref{fig:force_field}).
The inferred viscous coefficient is largest for damped oscillation ($0.68$) and deceleration ($0.63$) and smallest for uniform motion ($0.14$) and size change ($0.12$) (Table~\ref{tab:dissipation}).
The physics code separates motion types better than the code of every other transition, with a linear-probe accuracy of $0.769$ (Table~\ref{tab:vis_eta}), although motion labels are never used to train the model.
For pendulums, even though six context frames of monocular pixel coordinates cover only part of a swing and carry no metric scale, the learned squared frequency still correlates positively with $g/L$ ($\rho=0.54$, $95\%$ CI $[0.31, 0.72]$, Table~\ref{tab:param_recovery}), and LaWM attains the lowest error when the test pendulum lengths lie outside the training range ($0.386$, Table~\ref{tab:phys_extrap}).

\subsection{Ablations}
\label{app:ablation}

Table~\ref{tab:ablation} removes one component of LaWM per row, with perception fixed to prompted tracking, and reports the mean and standard deviation over $5$ seeds. In the standard setting, removing the forces or the context inference changes the mean error by less than $0.01$. Removing both (VIN) raises it to $0.420$. Removing the context smoothing gives the same mean with a much larger spread over seeds ($\pm0.092$), while its median ($0.275$) stays close to that of the full model. Without the forces the divergence rate is slightly lower ($5.2\%$ against $5.7\%$), but the median rises from $0.266$ to $0.274$. Table~\ref{tab:ablation_full} gives the per-motion results.

\begin{table}[H]
\centering
\small
\setlength{\tabcolsep}{3pt}
\caption{Ablation of LaWM components.}
\label{tab:ablation}
\resizebox{\ifdim\width>\columnwidth\columnwidth\else\width\fi}{!}{%
\begin{tabular}{lccc}
\toprule
Variant & Mean & Median & Div.\ (\%) \\
\midrule
Full LaWM & \bfseries \best{0.368$\pm$0.012} & \bfseries \best{0.266} & \bfseries \secondbest{5.7} \\
w/o forces & \secondbest{0.371$\pm$0.013} & \secondbest{0.274} & \best{5.2} \\
w/o context inference (FVIN) & 0.377$\pm$0.007 & 0.296 & 6.2 \\
w/o both (VIN) & 0.420$\pm$0.011 & 0.338 & 7.2 \\
w/o context smoothing & 0.420$\pm$0.092 & 0.275 & 5.8 \\
\bottomrule
\end{tabular}}
\end{table}

\begin{table*}[!t]
\centering
\small
\setlength{\tabcolsep}{4pt}
\caption{Per-motion results of the LaWM ablations (full version of Table~\ref{tab:ablation}).}
\label{tab:ablation_full}
\resizebox{\ifdim\width>\textwidth\textwidth\else\width\fi}{!}{%
\begin{tabular}{lccccccc}
\toprule
Motion & w/o forces & w/o context inference (FVIN) & w/o both (VIN) & w/o context smoothing & + gyroscopic forces & + polynomial physics code & Full LaWM \\
\midrule
Uniform & 0.343 & \secondbest{0.341} & 0.384 & 0.423 & 0.384 & \best{0.316} & \bfseries 0.383 \\
Acceleration & 0.312 & 0.334 & 0.381 & 0.347 & \secondbest{0.311} & 0.322 & \bfseries \best{0.308} \\
Deceleration & 0.384 & \secondbest{0.299} & 0.306 & 0.442 & 0.354 & 0.321 & \bfseries \best{0.296} \\
Parabolic & 0.388 & \secondbest{0.376} & 0.391 & 0.415 & 0.394 & 0.378 & \bfseries \best{0.374} \\
Parabolic + rotation & 0.335 & \secondbest{0.331} & 0.351 & 0.362 & 0.332 & 0.365 & \bfseries \best{0.328} \\
3D & 0.244 & \secondbest{0.240} & 0.267 & 0.312 & \secondbest{0.240} & 0.256 & \bfseries \best{0.214} \\
Size change & \best{0.530} & 0.552 & 0.570 & 0.593 & 0.566 & 0.556 & \bfseries \secondbest{0.549} \\
Slope & 0.264 & 0.279 & 0.312 & 0.305 & \best{0.244} & 0.284 & \bfseries \secondbest{0.259} \\
Circular & \best{0.376} & 0.441 & 0.420 & 0.481 & 0.392 & 0.396 & \bfseries \secondbest{0.389} \\
Rotation & 0.351 & 0.357 & 0.413 & 0.382 & \secondbest{0.348} & 0.352 & \bfseries \best{0.344} \\
Damped oscillation & 0.384 & 0.391 & 0.491 & 0.426 & \best{0.344} & 0.386 & \bfseries \secondbest{0.375} \\
Deformation & 0.308 & \secondbest{0.239} & 0.338 & 0.264 & 0.246 & 0.282 & \bfseries \best{0.236} \\
\midrule
Mean & 0.371$\pm$0.013 & 0.377$\pm$0.007 & 0.420$\pm$0.011 & 0.420$\pm$0.092 & \secondbest{0.370$\pm$0.009} & 0.373$\pm$0.011 & \bfseries \best{0.368$\pm$0.012} \\
Median & 0.274 & 0.296 & 0.338 & 0.275 & \secondbest{0.270} & 0.280 & \bfseries \best{0.266} \\
Div.\ (\%) & \best{5.2} & 6.2 & 7.2 & 5.8 & 5.8 & 6.0 & \bfseries \secondbest{5.7} \\
\bottomrule
\end{tabular}}
\end{table*}

In Table~\ref{tab:ablation_full}, each column changes one component, with the mean over $5$ seeds. The last two columns add components that the final model does not use. With gyroscopic forces the mean error is $0.370\pm0.009$ (median $0.270$, Div.\ $5.8\%$), and with the polynomial physics code it is $0.373\pm0.011$ (median $0.280$, Div.\ $6.0\%$). Both are within the seed variance of the full model.
Table~\ref{tab:ablation_hard} repeats the ablations beyond the standard setting with the same tracked coordinates. The $64$-frame column is the error up to frame $64$ on the $128$-frame IRIS windows of Fig.~\ref{fig:horizon_curve}. Here the components matter much more. Without the forces, the $42$-frame error rises from $0.247$ to $0.573$ and the $64$-frame error from $0.612$ to $1.405$. Without the context inference (FVIN), the $64$-frame error reaches $1.706$ and the error with $10\%$ of the data reaches $0.833$. VIN, which drops both, is the most accurate on unseen motion types ($0.375$ against $0.436$) but the least stable over long horizons ($2.154$ at $64$ frames).

\begin{table}[H]
\centering
\small
\setlength{\tabcolsep}{3pt}
\caption{Ablations beyond the standard setting.}
\label{tab:ablation_hard}
\resizebox{\ifdim\width>\columnwidth\columnwidth\else\width\fi}{!}{%
\begin{tabular}{lcccc}
\toprule
Variant & 42 frames & 64 frames & Unseen types & 10\% data \\
\midrule
w/o forces & 0.573 & 1.405 & 0.446 & 0.544 \\
w/o context inference (FVIN) & \secondbest{0.287} & 1.706 & \secondbest{0.434} & 0.833 \\
w/o both (VIN) & 0.596 & 2.154 & \best{0.375} & \secondbest{0.511} \\
w/o context smoothing & 0.324 & \secondbest{0.679} & 0.440 & 0.616 \\
Full LaWM & \bfseries \best{0.247} & \bfseries \best{0.612} & \bfseries 0.436 & \bfseries \best{0.504} \\
\bottomrule
\end{tabular}}
\end{table}

\subsection{Recovery of Measured Physical Parameters}
\label{app:param_recovery}

Table~\ref{tab:param_recovery} compares quantities implied by the learned Lagrangian with measurements that are never used in training, using the Spearman correlation and $95\%$ bootstrap confidence intervals over clips. Only the ordering is compared, since the absolute scale of these quantities is not recovered. Evaluated over the smoothed rollout trajectories, the conservative acceleration on ramps correlates strongly with the sine of the slope angle ($\rho=0.72$), and the squared frequency on pendulums exhibits a clear positive correlation with $g/L$ ($\rho=0.54$), confirming the physical ordering directly from visual observations without supervision.
Table~\ref{tab:dissipation} lists the inferred viscous coefficient per motion type as the median over test clips under normalized per-clip aggregation, with the mean and standard deviation over $5$ seeds. The coefficients are never supervised, yet their ordering is cleanly separated and matches the physical hierarchy in Fig.~1 and Sec.~4.4 of the main paper: damped oscillation and deceleration, where friction dominates the dynamics, yield the largest coefficients ($0.68$ and $0.63$), whereas uniform motion and size change yield the smallest ($0.14$ and $0.12$).

\begin{table}[H]
\centering
\small
\setlength{\tabcolsep}{3pt}
\caption{Correlation between learned and measured physical quantities.}
\label{tab:param_recovery}
\resizebox{\ifdim\width>\columnwidth\columnwidth\else\width\fi}{!}{%
\begin{tabular}{lcccc}
\toprule
Measured quantity & Clips & Learned quantity & Spearman $\rho$ & 95\% CI \\
\midrule
Slope angle ($\sin\theta$) & 47 & Conservative acceleration & 0.72 & [0.56, 0.84] \\
Pendulum length ($g/L$) & 48 & Squared frequency & 0.54 & [0.31, 0.72] \\
\bottomrule
\end{tabular}}
\end{table}

\begin{table}[H]
\centering
\small
\setlength{\tabcolsep}{4pt}
\caption{Inferred viscous coefficient $\|c(\eta)\|$ per motion type.}
\label{tab:dissipation}
\begin{tabular}{lc}
\toprule
Motion & $\|c(\eta)\|$ \\
\midrule
Damped oscillation & 0.68$\pm$0.09 \\
Deceleration & 0.63$\pm$0.10 \\
Deformation & 0.46$\pm$0.08 \\
Acceleration & 0.41$\pm$0.09 \\
Parabolic & 0.35$\pm$0.08 \\
Slope & 0.33$\pm$0.09 \\
3D & 0.27$\pm$0.07 \\
Rotation & 0.24$\pm$0.06 \\
Circular & 0.19$\pm$0.05 \\
Uniform & 0.14$\pm$0.04 \\
Size change & 0.12$\pm$0.03 \\
\bottomrule
\end{tabular}
\end{table}

\subsection{Learned Force Fields}
\label{app:force_field}

\begin{figure*}[!t]
\centering
\includegraphics[width=\textwidth]{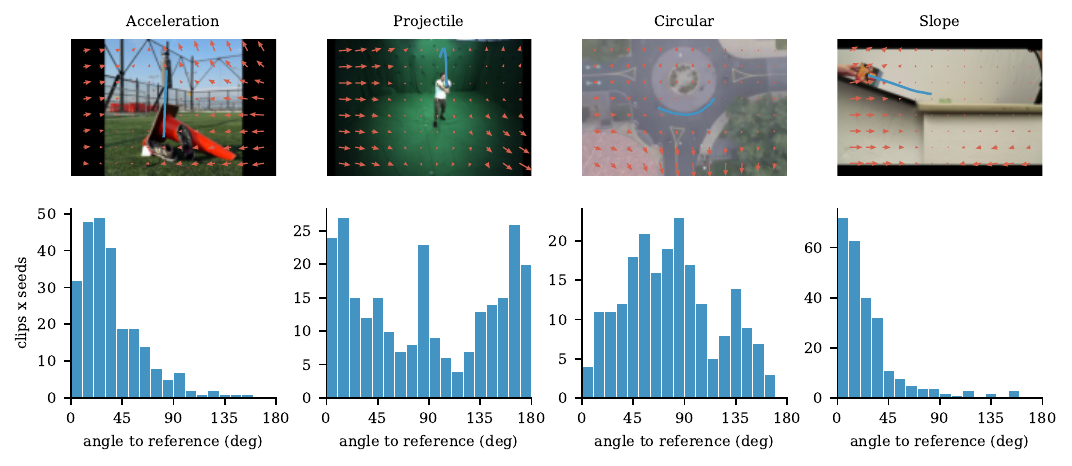}
\caption{Learned force fields on real test clips.}
\label{fig:force_field}
\end{figure*}

Figure~\ref{fig:force_field} shows the conservative acceleration $-M^{-1}\nabla V$ learned on real test clips. Arrows show this acceleration over the frame, and histograms show its angle to a reference direction, which is vertical for acceleration (falling) and projectile clips, toward the center for circular clips, and along the ramp for slope clips.
On \emph{Acceleration} and \emph{Slope} clips, the learned acceleration aligns closely with gravity and with the ramp direction ($<35^\circ$). On \emph{Projectile} and \emph{Circular} clips the angles spread more widely. In short monocular projectile clips, camera perspective and upward launch phases can partly confound the inferred potential gradient. In real circular scenes, such as vehicles on a roundabout, the motion is driven by thrust and steering rather than by a central conservative field, so the model assigns part of the dynamics to tangential and nonconservative components. The alignment on slope clips agrees with Table~\ref{tab:param_recovery}, where the conservative acceleration on ramps correlates strongly with the sine of the slope angle ($\rho=0.72$). Both results concern the conservative part of the model alone, since dissipation is carried by the separate nonconservative forces.

\subsection{Physics Code}
\label{app:vis_eta}
\label{app:eta_probe}

Table~\ref{tab:vis_eta} adds a visual object descriptor (frozen DINOv2 tokens of the prompted object, $64$ dimensions) to the input of the physics code encoder, for every transition that has a physics code. All runs use the same perception on the $554$ test clips with $5$ seeds, and Acc.\ is the accuracy of a linear motion-type classifier on the code. From the trajectory alone, the code of LaWM carries the most information about the motion type ($0.769$, against at most $0.665$ for the others). The visual descriptor raises every accuracy to between $0.763$ and $0.801$ and lowers the error of LaWM from $0.368$ to $0.335$, the lowest in the table. In error, the descriptor helps the GRU ($0.362$ to $0.338$) and the Transformer ($0.361$ to $0.337$) about as much, leaves the MLP unchanged ($0.355$), and makes the LNN slightly worse ($0.398$ to $0.408$). In accuracy, LaWM gains the least ($0.769$ to $0.801$, against gains of $0.10$ to $0.18$ for the others), which is consistent with its code already separating most motion types from the trajectory alone.
The probe measures how much of the motion type can be read linearly from the code, so a high accuracy means that clips of the same type receive similar codes even though the model never sees the label. Probe accuracy and prediction error are not ranked the same way. With the trajectory only, the MLP has the lowest error ($0.355$) but a lower accuracy than LaWM ($0.665$ against $0.769$), so the accuracy describes how the code is organized rather than how well the model predicts. With the visual descriptor, LaWM, the Transformer, and the GRU reach $0.335$, $0.337$, and $0.338$, within one standard deviation over seeds of one another. Without the descriptor, the Transformer has the least informative code ($0.606$).

\begin{table}[H]
\centering
\small
\setlength{\tabcolsep}{3pt}
\caption{Physics code with and without a visual object descriptor.}
\label{tab:vis_eta}
\begin{tabular}{lcccc}
\toprule
& \multicolumn{2}{c}{Trajectory only} & \multicolumn{2}{c}{+ visual} \\
\cmidrule(lr){2-3}\cmidrule(lr){4-5}
Transition & Mean & Acc. & Mean & Acc. \\
\midrule
MLP & \best{0.355$\pm$0.005} & \secondbest{0.665} & 0.355$\pm$0.012 & 0.763 \\
GRU & 0.362$\pm$0.004 & 0.652 & 0.338$\pm$0.004 & 0.782 \\
Transformer & \secondbest{0.361$\pm$0.014} & 0.606 & \secondbest{0.337$\pm$0.011} & 0.787 \\
Neural ODE & \secondbest{0.361$\pm$0.010} & 0.652 & 0.357$\pm$0.021 & 0.771 \\
LNN & 0.398$\pm$0.006 & 0.664 & 0.408$\pm$0.003 & \secondbest{0.796} \\
LaWM & \bfseries 0.368$\pm$0.012 & \bfseries \best{0.769} & \bfseries \best{0.335$\pm$0.009} & \bfseries \best{0.801} \\
\bottomrule
\end{tabular}
\end{table}

\begin{figure}[H]
\centering
\includegraphics[width=\columnwidth]{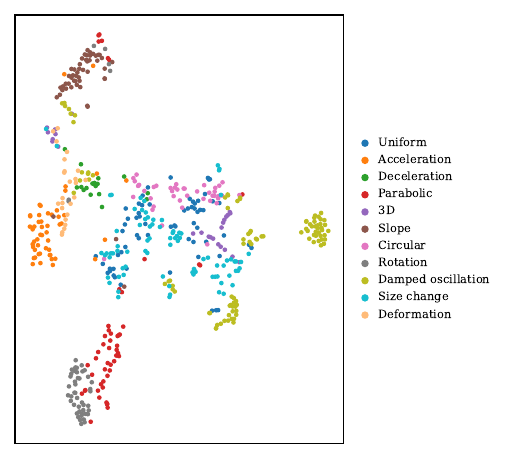}
\caption{2-D t-SNE embedding of the physics code $\eta$ on real test clips, colored by motion type. The motion label is never used to train the model.}
\label{fig:eta_tsne}
\end{figure}

The descriptor adds information about the object itself, such as whether it is a ball, a vehicle, or a pendulum, which six context frames of its trajectory may not reveal. The main results use the trajectory-only code (mean error $0.368$, as in Table~\ref{tab:same_perception_full}), so appearance can be added on top of them when visual features are available.
Figure~\ref{fig:eta_tsne} shows a t-SNE embedding of the physics code of LaWM on the real test clips. Each point is one clip, colored by its motion type, which the model never sees during training. Several motion types form separate groups, while others overlap near the center of the embedding, which agrees with a probe accuracy of $0.769$ rather than a perfect separation. Since t-SNE preserves local neighborhoods but not global distances, the figure shows which clips receive similar codes, not how far apart the groups are. The same code conditions the potential and the mass matrix of each clip (Appendix~\ref{app:impl}). Together with the viscous coefficients in Table~\ref{tab:dissipation} and the force fields in Fig.~\ref{fig:force_field}, the embedding shows that the inferred physics is organized by physical regime, although training uses neither motion labels nor measured physical quantities.

\subsection{Hyperparameter Sensitivity}
\label{app:sensitivity}

Table~\ref{tab:sensitivity} varies one hyperparameter at a time over $3$ seeds (seeds $0$ to $2$). The default is $\lambda_{\mathrm{DEL}}=0.1$, width $128$, and $\dim\eta=32$, and the solver iterations are in Table~\ref{tab:solver}. Every setting stays between $0.361$ and $0.375$ in mean error except $\lambda_{\mathrm{DEL}}=0$, which raises the error to $0.414$ and the spread over seeds to $\pm0.034$. The DEL loss is therefore needed, but its weight can vary over more than an order of magnitude without changing the result.

\begin{table}[H]
\centering
\small
\setlength{\tabcolsep}{4pt}
\caption{Hyperparameter sensitivity of LaWM.}
\label{tab:sensitivity}
\begin{tabular}{lccc}
\toprule
Setting & Mean & Median & Div.\ (\%) \\
\midrule
Default & \bfseries 0.362$\pm$0.010 & \bfseries 0.264 & \bfseries 5.3 \\
$\lambda_{\mathrm{DEL}}=0$ & 0.414$\pm$0.034 & 0.271 & 6.5 \\
$\lambda_{\mathrm{DEL}}=0.03$ & 0.367$\pm$0.009 & 0.270 & 4.5 \\
$\lambda_{\mathrm{DEL}}=0.3$ & 0.366$\pm$0.010 & 0.264 & 5.2 \\
$\lambda_{\mathrm{DEL}}=1$ & 0.370$\pm$0.004 & 0.274 & 5.2 \\
Width $64$ & 0.366$\pm$0.004 & 0.269 & 5.2 \\
Width $256$ & 0.375$\pm$0.014 & 0.273 & 4.7 \\
$\dim\eta=8$ & 0.367$\pm$0.003 & 0.276 & 5.0 \\
$\dim\eta=64$ & 0.361$\pm$0.008 & 0.261 & 5.3 \\
\bottomrule
\end{tabular}
\end{table}

\section{Perception}
\label{app:perception}

Perception enters LaWM as an exchangeable module.
Our prompted tracker reads generalized coordinates from the frozen DINOv2 features with no additional parameters, compared with $25.4$M for CoTracker3 and $224.4$M for SAM2, and every transition in Table~\ref{tab:trackers} is retrained on each source of coordinates, so the transitions are always compared on identical inputs.
With more accurate coordinates, the complete LaWM system improves from $0.368$ to $0.231$ with CoTracker3, LaWM attains the lowest mean error with CoTracker3, SAM2, and annotated coordinates, and its margin over FVIN, which shares the integrator but uses one Lagrangian for all clips, widens to $0.101$ with CoTracker3 and $0.063$ with annotated coordinates.
Inferring the physics per clip therefore becomes more valuable as perception improves, and the method benefits directly from advances in tracking.
The same tracker also decodes the predictions of DINO-WM and of the video generation models, so all of these comparisons share one readout (Tables~\ref{tab:dinowm} and~\ref{tab:videogen}).

\subsection{DINO-WM Baseline}
\label{app:dinowm}

In Table~\ref{tab:dinowm}, the DINO-WM predictor (depth $6$, $16$ heads, $14.0$M parameters) is trained for $4000$ steps with batch size $64$ on frozen DINOv2-L layer-20 tokens pooled to an $8\times12$ grid, over $5$ seeds. Decoded with the global readout of the baselines, DINO-WM reaches $0.949$, and our prompted tracking on its predicted features lowers this to $0.681$. The third row applies the same tracking to the real future features and gives $0.589$, which bounds the accuracy attainable at this token resolution. LaWM reaches $0.368$, well below this bound. The divergence rate follows the same order, from $37.0\%$ with the global readout to $24.0\%$ with prompted tracking and $22.0\%$ for the bound, against $5.7\%$ for LaWM. Even perfectly predicted features would therefore leave DINO-WM far behind at this resolution, so the limit lies in the coarse $8\times12$ token grid rather than in the predictor.

\begin{table}[H]
\centering
\small
\setlength{\tabcolsep}{3.5pt}
\caption{DINO-WM on the $554$ test clips.}
\label{tab:dinowm}
\resizebox{\columnwidth}{!}{%
\begin{tabular}{lccc}
\toprule
Decoder & Mean & Median & Div.\ (\%) \\
\midrule
Global readout (as baseline) & 0.949$\pm$0.006 & 0.789 & 37.0 \\
Prompted tracking on predicted features & 0.681$\pm$0.003 & 0.590 & 24.0 \\
\midrule
Tracking on real future features (bound) & \secondbest{0.589} & \secondbest{0.386} & \secondbest{22.0} \\
LaWM (ours) & \bfseries \best{0.368$\pm$0.012} & \bfseries \best{0.266} & \bfseries \best{5.7} \\
\bottomrule
\end{tabular}}
\end{table}

\subsection{Off-the-Shelf Trackers}\label{app:trackers}

\begin{table*}[!t]
\centering
\small
\setlength{\tabcolsep}{4pt}
\caption{Perception $\times$ transition grid.}
\label{tab:trackers}
\begin{tabular}{lcccccc}
\toprule
Perception & Extra params (M) & Loc.\ (px) & MLP & Transformer & FVIN & LaWM (ours) \\
\midrule
Global features & 0.005 & 89.5 & 4.639$\pm$7.732 & \secondbest{0.742$\pm$0.020} & 0.748$\pm$0.015 & \bfseries \best{0.730$\pm$0.004} \\
\quad + first-frame correction & 0.005 & 61.0 & 0.696$\pm$0.055 & \best{0.615$\pm$0.034} & 0.647$\pm$0.011 & \bfseries \secondbest{0.628$\pm$0.012} \\
Learned head & 0.75 & 72.0 & 0.656$\pm$0.013 & \secondbest{0.637$\pm$0.017} & 0.656$\pm$0.011 & \bfseries \best{0.603$\pm$0.005} \\
CoTracker3 & 25.4 & 7.0 & 0.248$\pm$0.007 & \secondbest{0.233$\pm$0.025} & 0.332$\pm$0.045 & \bfseries \best{0.231$\pm$0.007} \\
SAM2 (annotated subset) & 224.4 & 0.8 & 0.184$\pm$0.009 & \secondbest{0.183$\pm$0.019} & 0.205$\pm$0.007 & \bfseries \best{0.176$\pm$0.009} \\
Prompted tracking (ours) & 0 & \bfseries 19.4 & \best{0.355$\pm$0.005} & \secondbest{0.361$\pm$0.014} & 0.377$\pm$0.007 & \bfseries 0.368$\pm$0.012 \\
\midrule
Annotated coordinates (oracle) & -- & 0 & \secondbest{0.203$\pm$0.003} & 0.207$\pm$0.020 & 0.264$\pm$0.008 & \bfseries \best{0.201$\pm$0.007} \\
\bottomrule
\end{tabular}
\end{table*}

Table~\ref{tab:trackers} changes the source of the context coordinates and retrains every transition on each source, with the mean and standard deviation over $5$ seeds. Extra params counts the parameters beyond the shared frozen backbone, and Loc.\ is the localization error in pixels on the context frames. SAM2 is evaluated only on clips whose targets come from human annotations, since SAM2 masks provide the targets of the other clips. The colors compare the transitions within each row.
Better perception lowers the error of every transition, and the complete LaWM system improves from $0.368$ with prompted tracking to $0.231$ with CoTracker3, against $0.615$ for the best conventional system of Table~1 of the main paper.
With CoTracker3, annotated coordinates, and SAM2, LaWM is significantly more accurate than FVIN, which shares its forced variational integrator but uses one Lagrangian for all clips. The $95\%$ paired-bootstrap intervals of the difference are $[0.053, 0.161]$ with CoTracker3, $[0.048, 0.078]$ with annotated coordinates, and $[0.014, 0.045]$ with SAM2.
The gap to FVIN grows with the accuracy of the coordinates. It is $0.009$ and not significant with prompted tracking, and it reaches $0.101$ with CoTracker3, $0.063$ with annotated coordinates, and $0.029$ on the annotated subset with SAM2. Inferring the physics per clip thus matters more once perception is no longer the bottleneck.
Against the black-box transitions, LaWM attains the lowest mean with CoTracker3, SAM2, and annotated coordinates. It is significantly more accurate than the MLP with CoTracker3 ($[0.007, 0.027]$) and on par with or better than the Transformer in every row, while with prompted tracking the MLP is ahead by $0.013$.
The ranking against the black-box transitions therefore depends on the tracker, but for every tracker LaWM is within their range or better, and its mean error is below that of FVIN in every row. The absolute accuracy of the complete system is set by the tracker, which is an exchangeable module of the method.

\subsection{Video Generation Models}
\label{app:videogen}

In Table~\ref{tab:videogen}, open video generation models continue the same test clips from the context frames, and our prompted tracker follows the object in the generated frames. We report the error and the residual of the tracked trajectory with respect to the closed-form law of its motion type. The generated motion follows such a law about as closely as LaWM does (residuals from $0.017$ to $0.021$), yet the error of both video models stays near $0.6$, against $0.368$ for LaWM. The video models thus produce lawful motion that does not match the actual continuation of the clip.

\begin{table}[H]
\centering
\small
\setlength{\tabcolsep}{3pt}
\caption{Video generation models as world models.}
\label{tab:videogen}
\resizebox{\ifdim\width>\columnwidth\columnwidth\else\width\fi}{!}{%
\begin{tabular}{lcccc}
\toprule
Model & Mean & Median & Div.\ (\%) & Law residual \\
\midrule
LTX-Video 0.9.5 & 0.602 & 0.497 & \secondbest{17.0} & \best{0.017} \\
CogVideoX-5B-I2V (100 clips) & \secondbest{0.584} & \secondbest{0.495} & 18.0 & \secondbest{0.020} \\
LaWM (ours) & \bfseries \best{0.368} & \bfseries \best{0.266} & \bfseries \best{5.7} & \bfseries 0.021 \\
\bottomrule
\end{tabular}}
\end{table}

\section{Simulated Manipulation}\label{app:robot}
\subsection{Protocol}\label{app:robot_protocol}

\begin{table*}[!t]
\centering
\small
\setlength{\tabcolsep}{4pt}
\caption{Scenes, data, and evaluation protocol of the action-conditioned experiments with a Franka Panda arm in LIBERO kitchen scenes.}
\label{tab:robot_protocol}
\resizebox{\ifdim\width>\textwidth\textwidth\else\width\fi}{!}{%
\begin{tabular}{lcccccc}
\toprule
Scene & Episodes (train/val/test) & Mass range & Friction range & State source & Action & Success criterion \\
\midrule
Object pushing & 1795 / 283 / 457 & 0.1--0.6\,kg; unseen 0.6--1.0 & 0.2--0.8; unseen 0.8--1.1 & object mask & fingertip position, velocity & within $3$\,cm of the goal \\
Rope dragging & 2255 / 357 / 487 & 0.03--0.12\,kg & 0.2--0.8; unseen 0.8--1.0 & rope mask & fingertip position, velocity & prediction only \\
T-push & 4823 / 582 / 965 & 0.15--0.5\,kg; unseen 0.5--0.8 & 0.2--0.6; unseen 0.6--0.9 & simulator pose & fingertip position, displacement & within $2$\,cm and $15^\circ$ after $80$ steps \\
\bottomrule
\end{tabular}}
\end{table*}

External inputs enter the forced discrete Euler--Lagrange update through the same channel $F_k^{\mathrm{ext}} = B(\eta) u_k$ (Eq.~7) without modifying $B(\eta)$. In Table~\ref{tab:robot_full}, we compare two choices of the input vector $u_k$. \emph{Linear force} passes the raw fingertip action directly as $u_k$. \emph{Contact force} gates the fingertip action by the proximity of the end effector to the object, so that $u_k$ is nonzero only when the end effector is in contact with the manipulated object.

\subsection{Results}\label{app:robot_results}

\begin{figure*}[!t]
\centering
\includegraphics[width=0.95\textwidth]{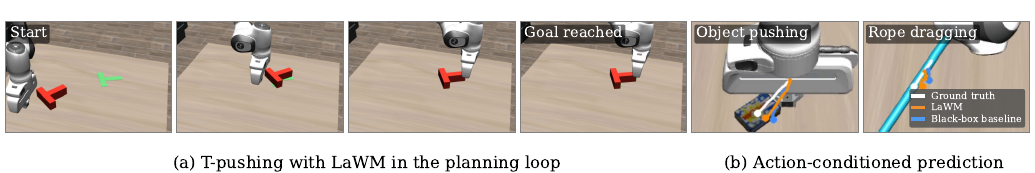}
\caption{Simulated manipulation with a Franka Panda arm in LIBERO kitchen scenes. (a) Closed-loop T-pushing with LaWM. (b) Predicted object-center trajectories for object pushing and rope dragging.}
\label{fig:robot}
\end{figure*}

In Fig.~\ref{fig:robot}(a), LaWM runs inside a model-predictive controller based on the cross-entropy method. The green outline marks the goal pose, and the episode shown is one in which LaWM succeeds. Fig.~\ref{fig:robot}(b) shows predicted trajectories of the object center for test clips of object pushing and rope dragging, with the first $6$ frames as context and reference lines from a black-box transition.
Table~\ref{tab:robot_full} reports the prediction error on the training distribution and the planning success rate (\%) on the training distribution and on unseen mass and friction (written as train / unseen), over $3$ seeds and all episodes. T-pushing uses $40$ episodes per test set and seed, and object pushing uses $60$. A dash marks a transition that was not evaluated on that scene.
With the contact force, LaWM has the lowest prediction error on all three scenes and the highest success rate in every planning column, reaching $34.2\%$ and $39.2\%$ on T-pushing and $31.7\%$ and $23.3\%$ on object pushing. With the linear force, its T-pushing success drops to $14.2\%$ and $12.5\%$, so the choice of $u_k$ has a large effect on planning.

\begin{table}[H]
\centering
\small
\setlength{\tabcolsep}{3pt}
\caption{Simulated manipulation, all transitions.}
\label{tab:robot_full}
\label{tab:robot}
\resizebox{\ifdim\width>\columnwidth\columnwidth\else\width\fi}{!}{%
\begin{tabular}{lccccc}
\toprule
& \multicolumn{3}{c}{Prediction} & \multicolumn{2}{c}{Planning success} \\
\cmidrule(lr){2-4}\cmidrule(lr){5-6}
Transition & Push & Rope & T-push & T-push & Push \\
\midrule
No action & -- & -- & -- & 0.0 / 0.0 & 0.0 / 0.0 \\
MLP (1st) & -- & 0.136 & 0.206 & 24.2 / \secondbest{37.5} & 27.8 / 16.7 \\
GRU & \secondbest{0.138} & 0.137 & 0.233 & 11.7 / 15.0 & \secondbest{29.4} / 20.0 \\
Transformer & \secondbest{0.138} & \secondbest{0.119} & \secondbest{0.174} & \secondbest{31.7} / 31.7 & 26.1 / 18.9 \\
Neural ODE & 0.149 & -- & 0.317 & 9.2 / 13.3 & 20.0 / 17.8 \\
FVIN & -- & -- & 0.313 & 16.7 / 19.2 & -- \\
LaWM, linear force & -- & -- & \bfseries 0.315 & \bfseries 14.2 / 12.5 & \bfseries 26.1 / \secondbest{20.6} \\
LaWM, contact force & \bfseries \best{0.132} & \bfseries \best{0.115} & \bfseries \best{0.168} & \bfseries \best{34.2} / \best{39.2} & \bfseries \best{31.7} / \best{23.3} \\
\bottomrule
\end{tabular}}
\end{table}

\section{Additional Experimental Details}
\label{app:details}

\subsection{Datasets}
\label{app:datasets}

Table~\ref{tab:datasets} lists the number of $16$-frame clips per motion type and split, with the splits made by recording event. Across the eleven motion types evaluated in the main tables, the test split comprises $554$ clips. Adding the $7$ Falling Objects clips gives $561$ test clips and $1{,}616$ clips in total.

\begin{table}[H]
\centering
\small
\setlength{\tabcolsep}{2.5pt}
\caption{Real-video dataset, number of $16$-frame clips per motion type and split.}
\label{tab:datasets}
\resizebox{\ifdim\width>\columnwidth\columnwidth\else\width\fi}{!}{%
\begin{tabular}{llccc}
\toprule
Motion & Sources & Train & Val & Test \\
\midrule
Uniform & MOT17, SDD & 85 & 26 & 64 \\
Acceleration & PisaBench, Physics-IQ, Morpheus, Delfys75 & 85 & 22 & 50 \\
Deceleration & PhysVid & 30 & 6 & 18 \\
Parabolic & GraviCap, Shuttlecock, TbD-3D & 85 & 22 & 51 \\
Parabolic + rotation & Falling Objects & 60 & 14 & 7 \\
3D & TbD-3D, IRIS & 29 & 8 & 25 \\
Size change & UA-DETRAC, Delfys75 & 95 & 22 & 83 \\
Slope & Physics 101, Delfys75, IRIS, Morpheus & 90 & 21 & 50 \\
Circular & SDD & 46 & 18 & 42 \\
Rotation & GS-Physics, Physics-IQ & 90 & 18 & 50 \\
Damped oscillation & IRIS, Delfys75, Morpheus, Physics 101 & 95 & 38 & 97 \\
Deformation & PhysTwin & 29 & 21 & 24 \\
\midrule
Total & & 819 & 236 & 561 \\
\bottomrule
\end{tabular}}
\end{table}

\subsection{Implementation Details}
\label{app:impl}
\label{app:implementation}

Each frame is resized to $448\times672$ and encoded independently with a frozen DINOv2-L backbone using layer-20 patch tokens on the $32\times48$ grid. The prompted tracker uses $\gamma=0.03$, a $48$-pixel local search radius, and fusion weight $\alpha=0.75$ without any trainable parameters. Context coordinates over the first $C=6$ frames are smoothed via least-squares quadratic projection and normalized by training-set statistics. The physics-code encoder $g_\rho$ is an MLP producing $\eta \in \mathbb{R}^{32}$, which conditions the potential $V_\theta$ and positive diagonal mass matrix $M_\theta$ (default width $128$) via FiLM. Nonconservative friction gates are initialized to $0.01$ with $\epsilon=0.02$. Each rollout step runs $J=4$ differentiable mass-preconditioned Newton iterations with step cap $\Delta_{\max}=2$. Models are trained with $\lambda_{\mathrm{DEL}}=0.1$ and $\lambda_{\mathrm{mass}}=\lambda_{\mathrm{sparse}}=10^{-3}$ using balanced sampling across motion types, and the checkpoint with the lowest validation error is selected. All baselines share the exact same context, smoothing, loss, optimizer, and training budget.

\paragraph{Evaluation settings.}
All experiments use $C=6$ observed context frames. In the standard setting (Table~1 of the main paper and Table~\ref{tab:same_perception_full}), clips have $T=16$ frames ($10$ predicted frames). In temporal extrapolation (\emph{Time $5\to10$}, Table~\ref{tab:general_full}), models are trained on $T=11$ frames ($5$ predicted frames) and evaluated on $T=16$ frames ($10$ predicted frames). For $42$-frame long-horizon damped oscillation prediction (Tables~\ref{tab:general_full} and~\ref{tab:ablation_hard}), clips have $T=48$ frames ($42$ predicted frames). For the $64$-frame horizon curve on IRIS windows (Fig.~\ref{fig:horizon_curve}), models trained on $T=16$ clips are rolled out autoregressively to $T=64$ frames ($58$ predicted frames).

\subsection{Statistical Protocol}
\label{app:bootstrap}

All main and generalization experiments are evaluated over $5$ random seeds ($3$ seeds for solver convergence, hyperparameter sensitivity, and simulated manipulation). A difference between two methods is considered significant at the seed level when it exceeds twice the sum of their standard deviations across seeds. To assess clip-level significance, we additionally compute $95\%$ confidence intervals and $p$-values using a paired bootstrap over the test clips, both overall (Table~\ref{tab:overall_sig}) and with Holm--Bonferroni correction for per-motion multiple comparisons (Table~\ref{tab:per_type}). All evaluation metrics and statistical tests were fixed prior to testing, and the test set was never used for hyperparameter tuning or checkpoint selection.

\subsection{Inference Cost}
\label{app:timing}

Table~\ref{tab:timing} reports the parameters and the rollout time in milliseconds on one A100 GPU, for a single clip ($B{=}1$) and for the full test set in one batch ($B{=}554$), with $10$ and $42$ predicted frames. A black-box transition evaluates one network per step, whereas the physics-structured transitions differentiate a learned energy at every step and, for VIN, FVIN, and LaWM, solve the implicit discrete Euler--Lagrange equation, so their cost is compared within the family.

\begin{table}[H]
\centering
\footnotesize
\setlength{\tabcolsep}{2.5pt}
\renewcommand{\arraystretch}{0.88}
\caption{Parameters and rollout time (ms) on one A100 GPU.}
\label{tab:timing}
\resizebox{\ifdim\width>\columnwidth\columnwidth\else\width\fi}{!}{%
\begin{tabular}{lccccc}
\toprule
& & \multicolumn{2}{c}{10 frames} & \multicolumn{2}{c}{42 frames} \\
\cmidrule(lr){3-4}\cmidrule(lr){5-6}
Transition & Params & $B{=}1$ & $B{=}554$ & $B{=}1$ & $B{=}554$ \\
\midrule
\multicolumn{6}{l}{\emph{Black-box}} \\
MLP & 88k & 4.2 & 4.6 & 16.6 & 17.4 \\
MLP (1st) & 88k & 3.9 & 4.2 & 15.5 & 16.7 \\
GRU & 86k & 1.7 & 1.8 & 6.3 & 6.4 \\
Transformer & 307k & 8.2 & 10.8 & 33.4 & 63.4 \\
RSSM & 154k & 6.1 & 6.7 & 16.0 & 16.3 \\
\multicolumn{6}{l}{\emph{Continuous-time}} \\
Neural ODE & 88k & 17.6 & 17.7 & 67.8 & 71.2 \\
\multicolumn{6}{l}{\emph{Physics-structured}} \\
HNN & 88k & 32.3 & 32.1 & 131.0 & 117.9 \\
LNN & 97k & 227.9 & 236.6 & 963.9 & 988.3 \\
VIN & 97k & 135.2 & 140.7 & 574.6 & 593.5 \\
FVIN & 102k & 111.0 & 114.2 & 466.7 & 484.4 \\
LaWM (ours) & \bfseries 102k & \bfseries 112.0 & \bfseries 115.1 & \bfseries 489.9 & \bfseries 526.1 \\
\bottomrule
\end{tabular}}
\end{table}

Within the physics-structured family, LaWM costs about as much as FVIN, which uses the same integrator, and less than VIN and the LNN, so inferring the physics code per clip adds almost no cost.
The cost scales with the number of solver iterations $J$.
Measured in one session on an RTX 6000 Ada GPU, $J=2$ and $J=1$ reduce the rollout time of LaWM by factors of $1.9$ and $3.7$ relative to $J=4$, while the error changes by less than the seed variance ($0.367$ and $0.369$ against $0.362$, Table~\ref{tab:solver}).

\section*{LLM Usage Statement}
ChatGPT was used to assist with language editing and improving manuscript clarity.
The authors remain responsible for all scientific claims, citations, and experimental results.
For the supplementary material, we used Claude to help with its layout and with the text that describes the reported results.

\end{multicols}

\end{document}